# BnB-ADOPT:
# An Asynchronous Branch-and-Bound DCOP Algorithm

**William Yeoh**                                                    WYEOH@USC.EDU
*Computer Science Department,*
*University of Southern California,*
*Los Angeles, CA 90089, USA*

**Ariel Felner**                                                  FELNER@BGU.AC.IL
*Information Systems Engineering,*
*Deutsche Telekom Labs,*
*Ben-Gurion University of the Negev,*
*Beer-Sheva, 85104, Israel*

**Sven Koenig**                                                   SKOENIG@USC.EDU
*Computer Science Department,*
*University of Southern California,*
*Los Angeles, CA 90089, USA*

## Abstract

Distributed constraint optimization (DCOP) problems are a popular way of formulating and solving agent-coordination problems. A DCOP problem is a problem where several agents coordinate their values such that the sum of the resulting constraint costs is minimal. It is often desirable to solve DCOP problems with memory-bounded and asynchronous algorithms. We introduce Branch-and-Bound ADOPT (BnB-ADOPT), a memory-bounded asynchronous DCOP search algorithm that uses the message-passing and communication framework of ADOPT (Modi, Shen, Tambe, & Yokoo, 2005), a well known memory-bounded asynchronous DCOP search algorithm, but changes the search strategy of ADOPT from best-first search to depth-first branch-and-bound search. Our experimental results show that BnB-ADOPT finds cost-minimal solutions up to one order of magnitude faster than ADOPT for a variety of large DCOP problems and is as fast as NCBB, a memory-bounded synchronous DCOP search algorithm, for most of these DCOP problems. Additionally, it is often desirable to find bounded-error solutions for DCOP problems within a reasonable amount of time since finding cost-minimal solutions is NP-hard. The existing bounded-error approximation mechanism allows users only to specify an absolute error bound on the solution cost but a relative error bound is often more intuitive. Thus, we present two new bounded-error approximation mechanisms that allow for relative error bounds and implement them on top of BnB-ADOPT.

## 1. Introduction

A distributed constraint optimization (DCOP) problem consists of agents, each responsible for taking on (= assigning itself) a value from its finite domain of values. The agents coordinate their values, which are subject to constraints. Two agents are constrained if they share a constraint. Each constraint has an associated constraint cost, which depends on the values of the constrained agents. A (complete) solution is an assignment of values to all agents, and a partial solution is an assignment of values to a subset of agents. The solution cost of a (partial or complete) solution is the sum of the constraint costs of all constraints resulting from the given assignment of values to agents. Solving a DCOP problem optimally means finding a solution with minimal solution cost and is NP-hard (Modi et al., 2005).

Formulating agent-coordination problems as constraint optimization (COP) problems, a specific type of weighted constraint satisfaction problems (Schiex, Fargier, & Verfaillie, 1995; Bistarelli,

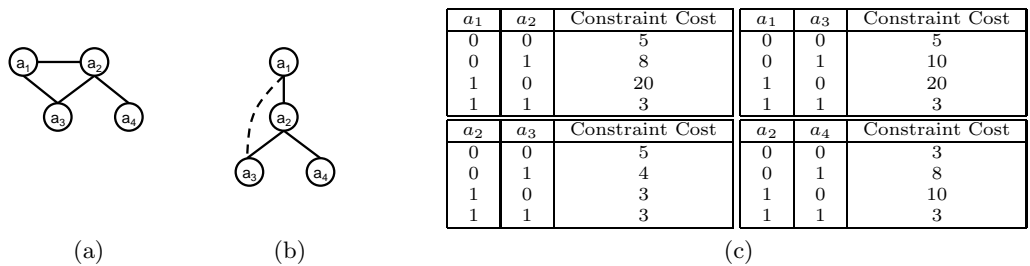

| $a_1$ | $a_2$ | Constraint Cost | | $a_1$ | $a_3$ | Constraint Cost |
|---|---|---|---|---|---|---|
| 0 | 0 | 5 | | 0 | 0 | 5 |
| 0 | 1 | 8 | | 0 | 1 | 10 |
| 1 | 0 | 20 | | 1 | 0 | 20 |
| 1 | 1 | 3 | | 1 | 1 | 3 |
| $a_2$ | $a_3$ | Constraint Cost | | $a_2$ | $a_4$ | Constraint Cost |
| 0 | 0 | 5 | | 0 | 0 | 3 |
| 0 | 1 | 4 | | 0 | 1 | 8 |
| 1 | 0 | 3 | | 1 | 0 | 10 |
| 1 | 1 | 3 | | 1 | 1 | 3 |

Figure 1: Example DCOP Problem

Montanari, Rossi, Schiex, Verfaillie, & Fargier, 1999), is more general than formulating them as the more common constraint satisfaction problems (Dechter, 2003). Constraint satisfaction problems have constraints that are either satisfied or unsatisfied. Solving a constraint satisfaction problem means finding a solution such that all constraints are satisfied. An example application is the scheduling of jobs in a job-shop, where constraints express that some jobs can only be performed by certain machines and some jobs can only be performed after some other jobs. There could potentially be multiple solutions that satisfy all constraints. However, some solutions might be more desirable than others. For example, one might prefer the solution with the shortest completion time. Unfortunately, constraint satisfaction problems cannot capture these preferences. However, COP problems are able to do so by using the constraint costs to represent the preferences.

DCOP algorithms are better suited compared to COP algorithms for problems that are naturally distributed. As a result, DCOP algorithms have been applied to coordinating unmanned aerial vehicles (Schurr, Okamoto, Maheswaran, Scerri, & Tambe, 2005), scheduling meetings (Maheswaran, Tambe, Bowring, Pearce, & Varakantham, 2004b; Petcu & Faltings, 2005b; Greenstadt, Grosz, & Smith, 2007; Zivan, 2008; Yeoh, Varakantham, & Koenig, 2009), coordinating sensor networks (Lesser, Ortiz, & Tambe, 2003; Zhang, Xing, Wang, & Wittenburg, 2003; Modi et al., 2005; Jain, Taylor, Tambe, & Yokoo, 2009; Stranders, Farinelli, Rogers, & Jennings, 2009; Zivan, Glinton, & Sycara, 2009), synchronizing traffic lights (Junges & Bazzan, 2008), planning truck routes (Ottens & Faltings, 2008) and managing power distribution networks (Kumar, Faltings, & Petcu, 2009).

It is common to visualize a DCOP problem as a constraint graph where the vertices are the agents and the edges are the constraints. Most DCOP algorithms operate on a pseudo-tree, which is a spanning tree of the (completely connected) constraint graph with the property that edges in the constraint graph connect a vertex with one of its ancestor or descendant vertices in the constraint tree (Freuder & Quinn, 1985; Bayardo & Miranker, 1995). An edge of the constraint graph that is not part of the pseudo-tree is a backedge. An agent $c$ is a pseudo-child agent of agent $p$ if agent $c$ is a descendant agent of agent $p$ in the pseudo-tree and they are constrained via a backedge. Similarly, agent $p$ is the pseudo-parent agent of agent $c$. Sibling subtrees represent independent DCOP subproblems (since no two agents in different sibling subtrees share a constraint). Figure 1(a) shows the constraint graph of an example DCOP problem with four agents that can each take on value 0 or value 1, Figure 1(b) shows one possible pseudo-tree where the assignments of values to agents $a_3$ and $a_4$ are independent DCOP subproblems (the dotted line is a backedge), and Figure 1(c) shows the constraint costs. For our example DCOP problem, a cost-minimal solution results if all agents take on value 1. The minimal solution cost is 12.

## 1.1 DCOP Algorithms

We now provide a taxonomy of DCOP algorithms. Figure 2 shows the taxonomy. DCOP algorithms are divided into two groups: *complete* and *incomplete* DCOP algorithms. Complete DCOP algorithms find cost-minimal solutions while incomplete DCOP algorithms are often faster but typically find suboptimal solutions.

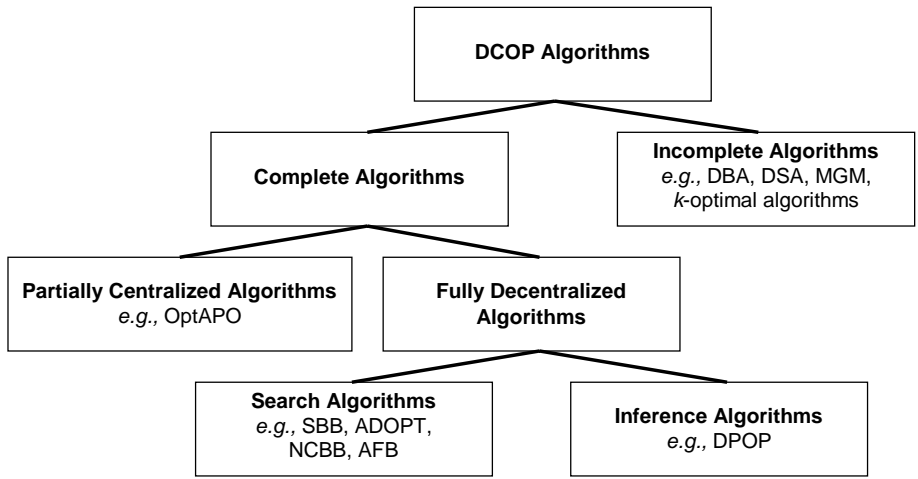

Figure 2: Taxonomy of DCOP Algorithms

### 1.1.1 Incomplete DCOP Algorithms

Incomplete DCOP algorithms typically use local search to find locally optimal solutions and can thus potentially get trapped in local minima. Nevertheless, since solving DCOP problems optimally is NP-hard, such DCOP algorithms are desirable for large DCOP problems where finding cost-minimal solutions might be slow. DBA (Yokoo & Hirayama, 1996), DSA (Fitzpatrick & Meertens, 2003), MGM (Maheswaran, Pearce, & Tambe, 2004a) and the more recent class of $k$-optimal DCOP algorithms (Pearce & Tambe, 2007; Bowring, Pearce, Portway, Jain, & Tambe, 2008; Greenstadt, 2009) are examples of incomplete DCOP algorithms.

### 1.1.2 Complete DCOP Algorithms

Complete DCOP algorithms are generally divided into two groups, namely *partially centralized* and *fully decentralized* DCOP algorithms.

#### Partially Centralized DCOP Algorithms

Partially centralized DCOP algorithms allow some agents to transfer their constraint information (= information regarding the constraints that they are involved in) to a central agent for processing. OptAPO (Mailler & Lesser, 2004) is an example of a partially centralized DCOP algorithm that uses cooperative mediation, where certain agents act as mediators to solve overlapping DCOP subproblems centrally.

#### Fully Decentralized DCOP Algorithms

Fully decentralized DCOP algorithms do not have central agents that collect constraint information of other agents that are not constrained with them. Rather, every agent has access to only its own constraint information. Fully decentralized DCOP algorithms are generally divided into two groups, namely DCOP *inference* and *search* algorithms.

- **DCOP inference algorithms:** DCOP inference algorithms typically use dynamic programming to propagate aggregated constraint costs from one agent to another agent and thus reduce the DCOP problem size by one agent at each step. They repeat this procedure until the DCOP problem size is reduced to only one agent and the solution space (= space of all possible partial solutions) thus cannot be reduced anymore. The sole remaining agent has then sufficient knowledge to find a cost-minimal solution. DPOP (Petcu & Faltings, 2005b) is an example of a DCOP inference algorithm. The number of messages sent between agents is only linear in the number of agents. However, its memory requirements are exponential in the induced width of the DCOP problem. The induced width depends on the number of backedges in the pseudo-tree. It can be as large as the number of agents minus one if the constraint graph is fully connected and every agent is thus constrained with every other agent.

| DCOP Algorithm | Search Strategy | Agent Operation | Communication | Agent Ordering |
|---|---|---|---|---|
| SBB | DFBnB | sequential & synchronous | point-to-point with neighbors | chain |
| ADOPT | best-first | concurrent & asynchronous | point-to-point with neighbors | tree |
| NCBB | DFBnB | sequential & synchronous | point-to-point with neighbors | tree |
| AFB | DFBnB | concurrent & asynchronous | broadcast to all agents | chain |
| BnB-ADOPT | DFBnB | concurrent & asynchronous | point-to-point with neighbors | tree |

Table 1: Properties of DCOP Search Algorithms

- **DCOP search algorithms:** DCOP search algorithms use search strategies to search through the solution space to find a cost-minimal solution. ADOPT (Modi et al., 2005) uses best-first search, and SBB (Hirayama & Yokoo, 1997), NCBB (Chechetka & Sycara, 2006), AFB (Gershman, Meisels, & Zivan, 2009) and our new DCOP search algorithm, BnB-ADOPT, use depth-first branch-and-bound search. Their memory requirements are only polynomial in the number of agents. However, the number of messages sent between agents can be exponential in the number of agents.

Therefore, both groups of fully decentralized DCOP algorithms are desirable under different conditions as there is a tradeoff between space (memory requirements) and time (number of messages sent).

## 1.2 Motivation

We now describe the motivation behind our work.

### 1.2.1 BnB-ADOPT

We study DCOP search algorithms because they can be memory-bounded. This property is important for applications, such as sensor networks, where every agent/sensor has only a fixed amount of memory available. As a result, several DCOP search algorithms, such as SBB, ADOPT, NCBB and AFB, were developed with this limitation in mind. As described earlier, their memory requirements are polynomial in the number of agents. Table 1 shows the properties of these DCOP search algorithms as well as the properties of our new DCOP search algorithm, BnB-ADOPT. We now describe each property in more detail and justify the properties of BnB-ADOPT.

- **Search strategy:** ADOPT uses best-first search to search the solution space, while SBB, NCBB and AFB use depth-first branch-and-bound (DFBnB) search. Best-first search repeatedly searches for the next best partial solution until it finds a cost-minimal solution. The next best partial solution is the cost-minimal partial solution among all partial solutions that have not yet been found. Depth-first branch-and-bound search starts by finding a complete (but often suboptimal) solution and stores its solution cost as the upper bound. It then continues to search for a solution whose solution cost is less than the upper bound. It stores the solution cost of this solution as the upper bound, and the search proceeds until it can no longer find a solution whose solution cost is less than the upper bound.

For centralized search, it is known that search problems with depth-bounded search trees can often be solved faster with depth-first branch-and-bound search than with memory-bounded best-first search because memory-bounded best-first search algorithms, such as RBFS (Korf, 1993), need to repeatedly reconstruct partial solutions that they purged from memory. Depth-first branch-and-bound search algorithms are memory-bounded but do not suffer from this problem (Zhang & Korf, 1995). Since DCOP problems are search problems with depth-bounded search trees, we hypothesize that depth-first branch-and-bound search might be faster than best-first search. Therefore, we decided that BnB-ADOPT should *use depth-first branch-and-bound search*.

- **Agent operation:** Agents of SBB and NCBB operate sequentially. Only agents with tokens are active while the other agents remain idle. Once the token-holding agents are done, they pass their tokens on and then remain idle. On the other hand, agents of ADOPT and AFB operate concurrently (= at all times). Agents that operate concurrently might be able to solve DCOP problems faster than agents that operate sequentially since the former agents can perform potentially useful computation instead of having to wait for other agents. Therefore, we decided that all agents of BnB-ADOPT should *operate concurrently*. Agents of SBB and NCBB also operate synchronously. Communication between agents is often in form of messages. Synchronous agents operate in cycles (Modi et al., 2005). A cycle is the time required for an agent to process all incoming messages in its queue and send all outgoing messages, which are then processed by the receiving agents in the next cycle (see Section 6.1 for more details). Therefore, all agents wait until the last agent is done sending its messages before they start a new cycle. On the other hand, asynchronous agents, such as agents of ADOPT and AFB, are able to operate independently of each other, which often increases robustness (Silaghi, Landwehr, & Larrosa, 2004). For example, all synchronous agents are affected if a single communication link suffers from congestion while only a small number of asynchronous agents are affected. We therefore decided that agents of BnB-ADOPT should *operate asynchronously*.

- **Communication:** DCOP search algorithms such as SBB, ADOPT and NCBB restrict communication to agents that share constraints. This restriction is motivated by applications such as sensor networks where communication is restricted to neighboring agents/sensors due to their limited communication radius. Neighboring sensors share constraints since they need to coordinate to sense the areas near them. DCOP search algorithms such as AFB do not have this restriction and allow agents to broadcast messages to all other agents. We decided that agents of BnB-ADOPT should obey the restrictions of applications such as sensor networks and thus *communicate only with neighboring agents*.

- **Agent ordering:** All DCOP search algorithms mentioned above start with a pre-processing step that arranges the agents into a pseudo-tree. DCOP search algorithms such as SBB and AFB arrange the agents into a chain, while ADOPT and NCBB arrange the agents into a tree. A tree ordering can capture independent DCOP subproblems (represented as sibling subtrees) while a chain ordering can not. DCOP search algorithms that operate on trees can thus operate on independent DCOP subproblems independently, while DCOP search algorithms that operate on chains can not. Therefore, we decided that BnB-ADOPT should *arrange agents into a tree*.

ADOPT has all preferred properties mentioned above except that it uses best-first search. We therefore introduce BnB-ADOPT, a memory-bounded asynchronous DCOP search algorithm that uses the message passing and communication framework of ADOPT but changes the search strategy of ADOPT from best-first search to depth-first branch-and-bound search.

### 1.2.2 Bounded-Error Approximations

Solving DCOP problems optimally is NP-hard, which makes it advantageous to allow users to trade off solution cost for a smaller runtime. It is also desirable to have the error of the resulting solution cost be bounded to provide guarantees on the solution cost. ADOPT is, to the best of our knowledge, the only DCOP search algorithm with this property. Its Absolute Error Mechanism allows its users to specify an absolute error bound on the solution cost, for example, that the solution cost should be at most 10 larger than the minimal solution cost. However, it is often much more desirable to specify a relative error bound on the solution cost, for example, that the solution cost should be at most 10 percent larger than the minimal solution cost or, equivalently, 1.1 times larger than the minimal solution cost. This cannot be done with the Absolute Error Mechanism without knowing the minimal solution cost a priori. Thus, we propose two approximation mechanisms that allow users to specify a relative error bound on the solution cost, namely the Relative Error Mechanism and the Weighted Heuristics Mechanism, and implement them on top of BnB-ADOPT. These approximation mechanisms allow BnB-ADOPT to find solutions with bounded errors faster than cost-minimal solutions.

### 1.3 Experimental Results

We experimentally compare ADOPT, BnB-ADOPT and NCBB on three different DCOP problem types, namely graph coloring problems, sensor network problems and meeting scheduling problems. Our results show that BnB-ADOPT is up to one order of magnitude faster (measured in the number of non-concurrent constraint checks and the number of cycles) than ADOPT on a variety of large DCOP problems. BnB-ADOPT can also be inferred to be faster than SBB since ADOPT is faster than SBB (Modi et al., 2005). BnB-ADOPT is also as fast as NCBB on most of these DCOP problems. Our results for the suboptimal variants of BnB-ADOPT show that the Weighted Heuristics Mechanism dominates both the Absolute Error Mechanism and Relative Error Mechanism.

### 1.4 Article Structure

This article is organized as follows: We formalize DCOP problems in Section 2 and describe our DCOP search algorithm, BnB-ADOPT, in Section 3. We describe approximation mechanisms that allow BnB-ADOPT to find solutions with bounded error in Section 4. We outline correctness and completeness proofs of BnB-ADOPT in Section 5. Lastly, we present our experimental evaluations in Section 6 and our conclusions in Section 7.

## 2. DCOP Problems

In this section, we formally define distributed constraint optimization (DCOP) problems and describe their solution space.

### 2.1 Definition of DCOP Problems

A DCOP problem is defined by the following elements:

- a finite set of agents $A = \{a_1, a_2, ..., a_n\}$;

- a set of finite domains $D = \{Dom(a_1), Dom(a_2), ..., Dom(a_n)\}$, where $Dom(a_i)$ is the domain of possible floating point values of agent $a_i \in A$; and

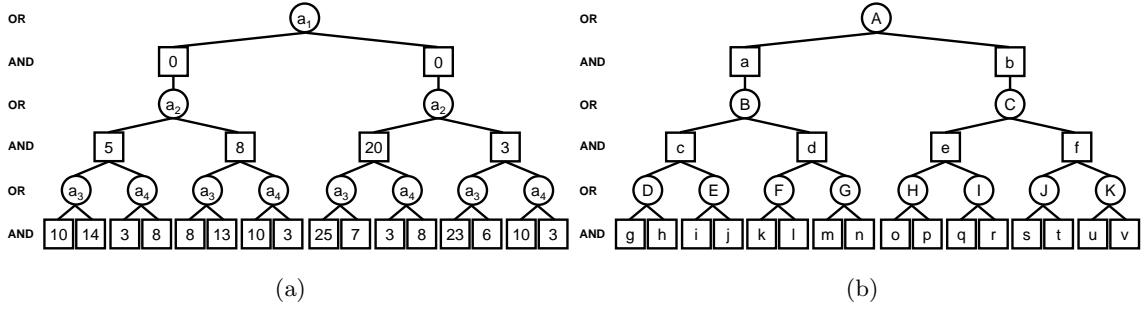

Figure 3: AND/OR Search Tree

- a set of binary constraints $F = \{f_1, f_2, ..., f_m\}$, where each constraint $f_i$: $Dom(a_{i_1}) \times Dom(a_{i_2}) \rightarrow \mathbb{R}^+ \cup \infty$, specifies its non-negative constraint cost as a function of the values of the distinct agents $a_{i_1}$ and $a_{i_2}$ that share the constraint.

The above definition assumes that each agent takes on one value rather than multiple values, for example, a different value for each constraint that it is involved in. These DCOP problems are more commonly formulated as each agent being responsible for the assignments of values to multiple variables. However, there exist techniques that reduce such DCOP problems to our DCOP problems (Burke & Brown, 2006). Thus, we use the terms agent and variable interchangeably. The above definition also assumes that constraints are binary (= between two agents) rather than $n$-ary (= between $n$ agents). One should be able to extend BnB-ADOPT to solve DCOP problems with $n$-ary constraints by using the same techniques that were proposed to extend ADOPT to solve DCOP problems with $n$-ary constraints (Modi et al., 2005). Additionally, we assume that the messages sent between agents can be delayed by a finite amount of time but are never lost.

## 2.2 Search Trees

The solution space of DCOP problems can be visualized with search trees. Traditional search trees or, synonymously, OR search trees (Marinescu & Dechter, 2009) assign values to agents sequentially. They do not utilize the fact that the values of agents that belong to independent DCOP subproblems do not have to be assigned sequentially. AND/OR search trees are based on pseudo-trees and remedy this issue (Marinescu & Dechter, 2009). Thus, we use AND/OR search trees and refer to them as search trees in this article. Their depth is bounded by (twice) the number of agents.

Figure 3(a) shows the search tree that is based on the pseudo-tree in Figure 1(b). Figure 3(b) labels each node of the search tree with an identifier to allow us to refer to the nodes easily. Circular nodes are OR nodes (labeled with upper-case letters) and correspond to agents. For example, the agent of node $C$ is agent $a_2$. Left branches of OR nodes correspond to the agents taking on value 0 and right branches correspond to the agents taking on value 1. Square nodes are AND nodes (labeled with lower-case letters) and correspond to the partial solutions from the root node to those nodes. For example, the partial solution of node $f$ is $\{(a_1, 1), (a_2, 1)\}$. The subtree rooted at an AND node represents the DCOP subproblem that assumes the partial solution of the AND node. For example, the subtree rooted at node $f$ represents the DCOP subproblem of assigning values to agents $a_3$ and $a_4$ given that $\{(a_1, 1), (a_2, 1)\}$. The number of independent DCOP subproblems within this DCOP subproblem is indicated by the number of branches exiting the AND node. For example, there are two branches exiting node $f$, indicating that there are two independent DCOP subproblems, namely of assigning values to agents $a_3$ and $a_4$. The numbers in the AND nodes are the delta costs of the nodes. The delta cost of an AND node is defined to be the sum of the constraint costs of all constraints in its partial solution that involve the agent of its parent OR node.

For example, the partial solution of node $v$ is $\{(a_1, 1), (a_2, 1), (a_4, 1)\}$. There are two constraints in this partial solution, namely the constraint between agents $a_1$ and $a_2$, which has constraint cost 3, and the constraint between agents $a_2$ and $a_4$, which also has constraint cost 3. Since the parent node of node $v$ is node $K$ with agent $a_4$, the delta cost of node $v$ is 3, namely the constraint cost of the latter constraint. The former constraint is not included since it does not involve agent $a_4$. The solution cost of a partial solution of an AND node is the sum of the delta costs of all AND nodes along the branch from the root node to that node. For example, the solution cost of the partial solution of node $v$ ($= 6$) is the sum of the delta costs of nodes $b$, $f$ and $v$. In our example DCOP problem, a cost-minimal solution is the union of the partial solutions of nodes $t$ and $v$ (all agents take on value 1). Thus, the minimal solution cost ($= 12$) is the sum of the delta costs of nodes $b$, $f$, $t$ and $v$.

## 3. BnB-ADOPT

In this section, we present Branch-and-Bound ADOPT (BnB-ADOPT). We do not describe BnB-ADOPT as a modification of ADOPT since this approach requires the readers to have an in-depth understanding of ADOPT. Instead, we give a stand-alone description of BnB-ADOPT that requires no knowledge of ADOPT, with the intention of creating a self-contained and hopefully easy-to-read description.

### 3.1 Search Strategies of ADOPT and BnB-ADOPT

We first describe centralized versions of the search strategies of ADOPT and BnB-ADOPT and omit technical details since these are described in more detail in later sections.

#### 3.1.1 SEARCH STRATEGY OF ADOPT

ADOPT (Modi et al., 2005) is a popular DCOP search algorithm (Modi & Ali, 2004; Ali, Koenig, & Tambe, 2005; Bowring, Tambe, & Yokoo, 2006; Davin & Modi, 2006; Pecora, Modi, & Scerri, 2006; Choxi & Modi, 2007; Silaghi & Yokoo, 2009; Matsui, Silaghi, Hirayama, Yokoo, & Matsuo, 2009) that traverses the search tree in a best-first search order. We now describe a simplified version of best-first search. The complete version can be found in (Marinescu & Dechter, 2007). Best-first search maintains a list that initially contains only the child AND nodes of the root node. It repeatedly performs the following operations: It expands the AND node with the smallest solution cost in the list by removing that node from the list and adding the grandchild AND nodes of that node into the list. For our example DCOP problem, best-first search expands the AND nodes in the search tree in Figure 3 for the first time in the following order, where the numbers in parentheses indicate the solution costs of the partial solutions of the expanded nodes: $a$ (0), $b$ (0), $f$ (3), $c$ (5), $v$ (6), $i$ (8), $d$ (8) and $t$ (9).

Figure 4 shows a simplified trace of ADOPT on our example DCOP problem. ADOPT terminates after fifteen steps with minimal solution cost 12. The numbers in the AND nodes are the delta costs of the nodes. The lower bound $LB^r_{X^r}$ is an optimistic estimate of the minimal solution cost. It is the smallest underestimated solution cost, over all solutions. The underestimated solution cost of a solution is the sum of the delta costs of all AND nodes of that solution whose parent OR node is the root node or whose grandparent AND node is expanded. For example, the underestimated solution cost of the solution $\{(a_1, 1), (a_2, 1), (a_3, 1), (a_4, 1)\}$ is 3 if node $b$ is expanded and nodes $f$, $t$ and $v$ are not expanded. The upper bound $UB^r_{X^r}$ is a pessimistic estimate of the minimal solution cost. It is the solution cost of the solution with the smallest solution cost found so far. ADOPT terminates when the upper bound $UB^r_{X^r}$ is no larger than the lower bound $LB^r_{X^r}$. In order to be memory-bounded, ADOPT maintains only one branch of the search tree (shaded grey in the figure) from the root node to the currently expanded node and thus needs to repeatedly reconstruct nodes that it purged from memory. For example, in Step 3, ADOPT has the branch to node $f$ in memory. The next node that best-first search expands is node $c$, and ADOPT discards the branch to node $f$ in Step 4. In Steps 6 and 7, it then needs to reconstruct the discarded branch to node $f$ in order to expand node $v$ in Step 8.

Figure 4: Trace of Simplified Memory-Bounded Best-First Search (Centralized ADOPT)

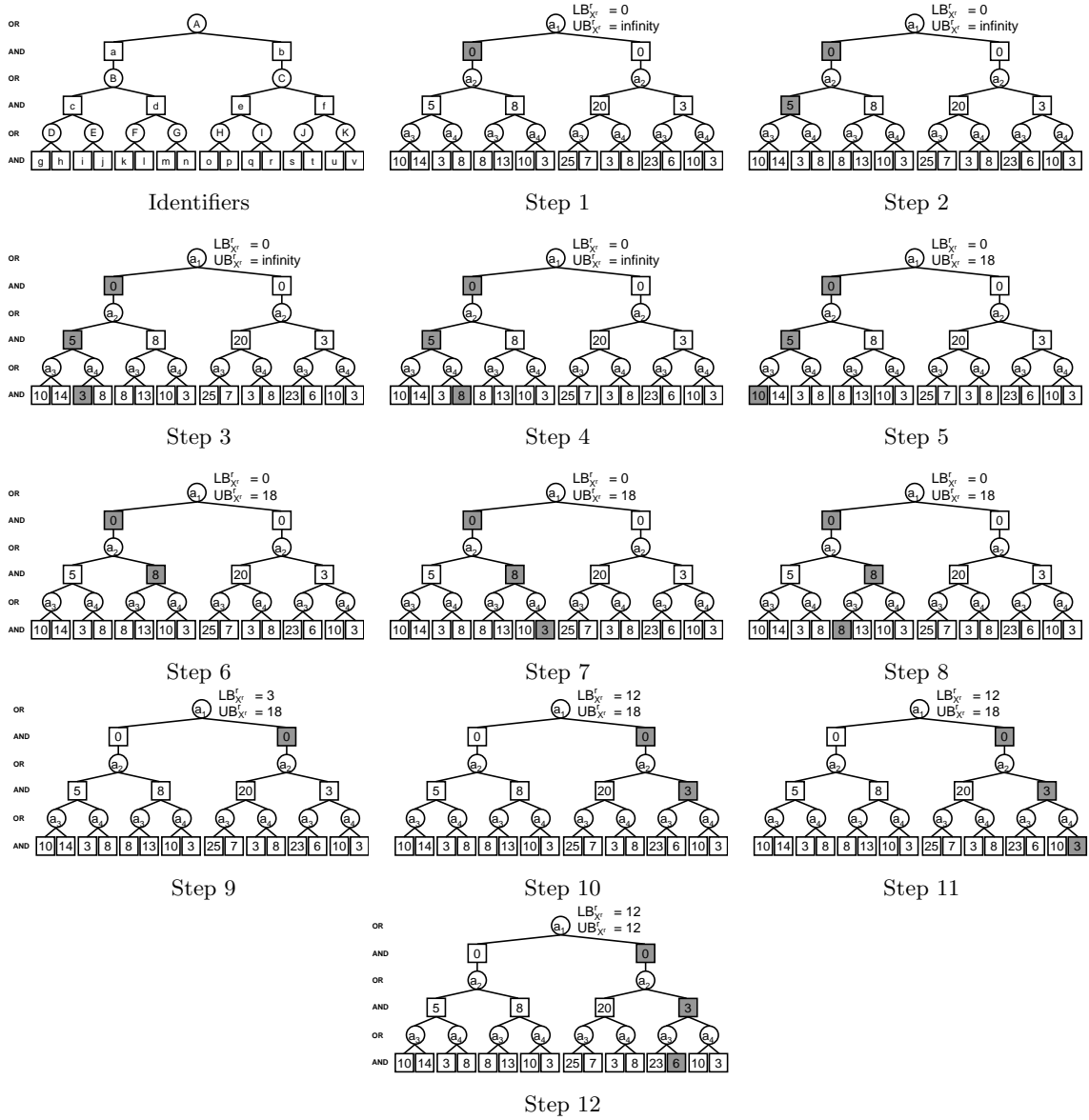

Figure 5: Trace of Simplified Depth-First Branch-and-Bound Search (Centralized BnB-ADOPT)

### 3.1.2 Search Strategy of BnB-ADOPT

We now describe a simplified version of depth-first branch-and-bound search. The complete version can be found in (Marinescu & Dechter, 2009). We use the same definitions of $LB_{X^r}^r$ and $UB_{X^r}^r$ as described earlier for Figure 4. Depth-first branch-and-bound search maintains a stack that initially contains only the child AND nodes of the root node. It expands the AND node on top of the stack by removing that node from the stack and performing the following check. If the solution cost of that node is no smaller than the upper bound $UB_{X^r}^r$, it prunes that node and repeats the operation. Otherwise, it adds the grandchild AND nodes of that node to the top of the stack and repeats the operation. It terminates when the upper bound $UB_{X^r}^r$ is no larger than lower bound $LB_{X^r}^r$. Depth-first branch-and-bound search can add the grandchild AND nodes of an expanded AND node (and the child AND nodes of the root node) in decreasing order of their solution costs instead of a random order to the top of the stack. This ordering ensures that depth-first branch-and-bound search expands the grandchild AND node with the smallest solution cost first. We use this improvement throughout the article. For our example DCOP problem, depth-first branch-and-bound search expands the AND nodes in the search tree in the following order, where it prunes the nodes in brackets: $a$ (0), $c$ (5), $i$ (8), $j$ (13), $g$ (15), [$h$ (19)], $d$ (8), $n$ (11), $k$ (16), [$m$ (18)], [$l$ (21)], $b$ (0), $f$ (3), $v$ (6) and $t$ (9). Figure 5 shows a trace of depth-first branch-and-bound search for our example DCOP problem. It is memory-bounded without having to repeatedly reconstruct nodes that it purged from memory but expands some nodes that a best-first search does not expand, such as node $j$ in Step 4. The depth-first branch-and-bound search terminates after twelve steps with minimal solution cost 12, which is three steps fewer than ADOPT.

## 3.2 Description of BnB-ADOPT

We now provide an incremental description of BnB-ADOPT. First, we provide the notations and key terms of BnB-ADOPT. Then, we describe how BnB-ADOPT updates its bounds, adheres to memory limitations, performs depth-first search and performs branch-and-bound. Finally, we introduce our enhanced final version of BnB-ADOPT and show both its pseudocode and its trace for our example DCOP problem.

### 3.2.1 NOTATION AND KEY TERMS

We adopt the following notation from ADOPT to describe BnB-ADOPT:

- $ValInit(a) \in Dom(a)$ is the initial value of agent $a \in A$;

- $CD(a) \subseteq A$ is the set of child and pseudo-child agents of agent $a \in A$;

- $C(a) \subseteq CD(a)$ is the set of child agents of agent $a \in A$;

- $pa(a) \in A$ is the parent agent of agent $a \in A$ except for the root agent;

- $P(a) \subseteq A$ is the set of ancestor agents (including the parent agent) of agent $a \in A$;

- $SCP(a) \subseteq P(a)$ is the set of ancestor agents (including the parent agent) of agent $a \in A$ that are parent or pseudo-parent agents of agent $a$ or one (or more) of its descendant agents; and

- $CP(a) \subseteq SCP(a)$ is the set of ancestor agents (including the parent agent) of agent $a \in A$ that are parent or pseudo-parent agents of agent $a$.

We adopt the following key terms from ADOPT to describe BnB-ADOPT:

- **Context ($X$):** The context $X^a$ of agent $a$ is the set of values of all ancestor agents of agent $a$. The context $X^r$ of the root agent $r$ is always equal to $\{\}$.

- **Delta cost ($\delta$):** The delta cost $\delta_{X^a}^a(d)$ is the sum of the constraint costs of all constraints that involve both agent $a$ and one of its ancestor agents, under the assumption that agent $a$ takes on value $d$ and its ancestor agents take on the values in context $X^a$. In the search tree, $\delta_{X^a}^a(d)$ is the delta cost of the AND node that has partial solution $X^a \cup (a, d)$. For example, $\delta_{\{(a_1,1)\}}^{a_2}(1)$ is the delta cost of node $f$ in Figure 3.

- **Gamma cost ($\gamma$):** The gamma costs $\gamma_{X^a}^a(d)$ and $\gamma_{X^a}^a$ are defined as follows:

$$\gamma_{X^a}^a(d) := \delta_{X^a}^a(d) + \sum_{c \in C(a)} \gamma_{X^a \cup (a,d)}^c \tag{1}$$

$$\gamma_{X^a}^a := \min_{d \in Dom(a)} \{\gamma_{X^a}^a(d)\} \tag{2}$$

for all agents $a$, all values $d$ and all contexts $X^a$. Thus, the gamma cost $\gamma_{X^a}^a(d)$ is the sum of the constraint costs of all constraints that involve agent $a$ or one of its descendant agents (that is, either both agent $a$ and one of its ancestor agents, both agent $a$ and one of its descendant agents, both a descendant agent and an ancestor agent of agent $a$ or two descendant agents of agent $a$) minimized over all possible values of its descendant agents, under the assumption that agent $a$ takes on value $d$ and its ancestor agents take on the values in context $X^a$. In the search tree, $\gamma_{X^a}^a(d)$ is the gamma cost of the AND node that has partial solution $X^a \cup (a, d)$. For example, $\gamma_{\{(a_1, 1)\}}^{a_2}(1)$ is the gamma cost of node $f$ in Figure 3. The gamma cost $\gamma_{X^a}^a$ is the sum of the constraint costs of all constraints that involve agent $a$ or one of its descendant agents minimized over all possible values of agent $a$ and its descendant agents, under the assumption that the ancestor agents of agent $a$ take on the values in context $X^a$. In the search tree, the gamma cost $\gamma_{X^a}^a$ is the gamma cost of the OR node whose agent is agent $a$ and whose parent AND node has partial solution $X^a$. For example, $\gamma_{\{(a_1, 1)\}}^{a_2}$ is the gamma cost of node $C$ in Figure 3. Therefore, the gamma cost of an AND node is the sum of its delta cost and the gamma costs of its child OR nodes, and the gamma cost of an OR node is the minimum of the gamma costs of its child AND nodes. For example, the gamma cost of node $f$ in Figure 3 is the sum of its delta cost and the gamma costs of nodes $J$ and $K$, and the gamma cost of node $C$ in Figure 3 is the minimum of the gamma costs of nodes $e$ and $f$.

Solving a DCOP problem optimally means to determine $\gamma_{X^r}^r$ for the root agent $r$ or, equivalently, the gamma cost of the root node since $\gamma_{X^r}^r$ is the minimal solution cost. It is not difficult for the agents to cache information that allows them to determine a cost-minimal solution.

### 3.2.2 Updating the Bounds

Every agent $a$ of BnB-ADOPT stores and updates several bounds on the gamma costs, namely $lb_{X^a}^{a,c}(d)$, $LB_{X^a}^a(d)$, $LB_{X^a}^a$, $ub_{X^a}^{a,c}(d)$, $UB_{X^a}^a(d)$ and $UB_{X^a}^a$ for all values $d$, all child agents $c$ and all contexts $X^a$, maintaining the following bound property:

$$LB_{X^a}^a \leq \qquad \gamma_{X^a}^a \qquad \leq UB_{X^a}^a \tag{3}$$

$$LB_{X^a}^a(d) \leq \qquad \gamma_{X^a}^a(d) \qquad \leq UB_{X^a}^a(d) \tag{4}$$

$$lb_{X^a}^{a,c}(d) \leq \qquad \gamma_{X^a \cup (a,d)}^c \qquad \leq ub_{X^a}^{a,c}(d) \tag{5}$$

In the search tree,

- $LB_{X^a}^a$ and $UB_{X^a}^a$ are lower and upper bounds, respectively, (on the gamma cost) of the OR node whose agent is agent $a$ and whose parent AND node has partial solution $X^a$;

- $LB_{X^a}^a(d)$ and $UB_{X^a}^a(d)$ are lower and upper bounds, respectively, (on the gamma cost) of the AND node that has partial solution $X^a \cup (a, d)$; and

- $lb_{X^a}^{a,c}(d)$ and $ub_{X^a}^{a,c}(d)$ are lower and upper bounds, respectively, (on the gamma cost) of the OR node whose agent is agent $c$ and whose parent AND node has partial solution $X^a \cup (a, d)$.

For example, $LB^{a_2}_{\{(a_1,1)\}}$ and $UB^{a_2}_{\{(a_1,1)\}}$ are bounds of node $C$ in Figure 3, $LB^{a_2}_{\{(a_1,1)\}}(1)$ and $UB^{a_2}_{\{(a_1,1)\}}(1)$ are bounds of node $f$, and $lb^{a_2,a_3}_{\{(a_1,1)\}}(1)$ and $ub^{a_2,a_3}_{\{(a_1,1)\}}(1)$ are bounds of node $J$. $lb^{a_2,a_3}_{\{(a_1,1)\}}(1)$, $ub^{a_2,a_3}_{\{(a_1,1)\}}(1)$, $LB^{a_3}_{\{(a_1,1),(a_2,1)\}}$ and $UB^{a_3}_{\{(a_1,1),(a_2,1)\}}$ are bounds of node $J$, but agent $a_2$ maintains the first two bounds while agent $a_3$ maintains the last two bounds.

Each agent $a$ uses the following update equations for all values $d$, all child agents $c$ and all contexts $X^a$ to initialize its bounds $lb^{a,c}_{X^a}(d)$ and $ub^{a,c}_{X^a}(d)$, where the heuristic values $h^{a,c}_{X^a}(d)$ are floating point numbers that are admissible and thus satisfy $0 \leq h^{a,c}_{X^a}(d) \leq \gamma^c_{X^a \cup (a,d)}$:

$$lb^{a,c}_{X^a}(d) := h^{a,c}_{X^a}(d) \tag{6}$$

$$ub^{a,c}_{X^a}(d) := \infty \tag{7}$$

Agent $a$ then uses repeatedly the following update equations for all values $d$, all child agents $c$, all contexts $X^a$ and all contexts $X^c$ ($= X^a \cup (a,d)$) to tighten the bounds:

$$lb^{a,c}_{X^a}(d) := \max\{lb^{a,c}_{X^a}(d), LB^c_{X^c}\} \tag{8}$$

$$LB^a_{X^a}(d) := \delta^a_{X^a}(d) + \sum_{c \in C(a)} lb^{a,c}_{X^a}(d) \tag{9}$$

$$LB^a_{X^a} := \min_{d \in Dom(a)}\{LB^a_{X^a}(d)\} \tag{10}$$

$$ub^{a,c}_{X^a}(d) := \min\{ub^{a,c}_{X^a}(d), UB^c_{X^c}\} \tag{11}$$

$$UB^a_{X^a}(d) := \delta^a_{X^a}(d) + \sum_{c \in C(a)} ub^{a,c}_{X^a}(d) \tag{12}$$

$$UB^a_{X^a} := \min_{d \in Dom(a)}\{UB^a_{X^a}(d)\} \tag{13}$$

The updates maintain the bound property and improve the bounds monotonically, that is, the lower bounds are monotonically non-decreasing and the upper bounds are monotonically non-increasing.[1] After a finite amount of time, $UB^a_{X^a} \leq LB^a_{X^a}$ for all agents $a$ and all contexts $X^a$. BnB-ADOPT terminates when its termination condition $UB^r_{X^r} \leq LB^r_{X^r}$ for the root agent $r$ is satisfied. Then, $UB^r_{X^r} \leq LB^r_{X^r}$ and the bound property $UB^r_{X^r} \geq LB^r_{X^r}$ together imply that $UB^r_{X^r} = \gamma^r_{X^r} = LB^r_{X^r}$, and the DCOP problem is solved optimally.

Figure 6 shows a simplified trace of the updates of the (lower and upper) bounds for our example DCOP problem. We assume that the updates proceed sequentially from the leaf agents to the root agent. Due to this simplification, the lower and upper bounds of each node are identical to its gamma cost and independent of the heuristic values. The numbers in the nodes are their bounds. Two agents maintain the bounds of OR nodes except for the root node. The figure shows the bounds that the parent agent maintains rather than the bounds that the child agent maintains. For example, the number in node $B$ is the bounds that agent $a_1$ rather than agent $a_2$ maintains. The bounds that the child agent maintains can be computed by taking the minimum of the bounds of the child AND nodes of the OR node. Agents update the bound of an AND node to the sum of its delta cost and the bounds of its child OR nodes according to update equations 9 and 12. They update the bound of an OR node to the minimum of the bounds of its child AND nodes according to update equations 10 and 13. A more detailed description of the trace is as follows:

- **Step 1:** Leaf agent $a_3$ updates the bounds of AND nodes $g$, $h$, $k$, $l$, $o$, $p$, $s$ and $t$ to their delta costs according to update equations 9 and 12 and the bounds of OR nodes $D$, $F$, $H$ and $J$ to the minimum of the bounds of their child AND nodes according to update equations 10 and 13. Similarly, leaf agent $a_4$ updates the bounds of AND nodes $i$, $j$, $m$, $n$, $q$, $r$, $u$ and $v$ to their delta costs according to update equations 9 and 12 and the bounds of OR nodes $E$, $G$, $I$ and $K$ to the minimum of the bounds of their child AND nodes according to update equations 10 and 13. The bounds of OR nodes $D$ to $K$ are not shown in the figure since they are not (yet) maintained by agent $a_2$.

---

1. Leaf agents use the same update equations. Since they do not have child agents, the sums over their child agents evaluate to 0. For example, $LB^a_{X^a}(d) = UB^a_{X^a}(d) = \delta^a_{X^a}(d)$ for all leaf agents $a$, all values $d$ and all contexts $X^a$.

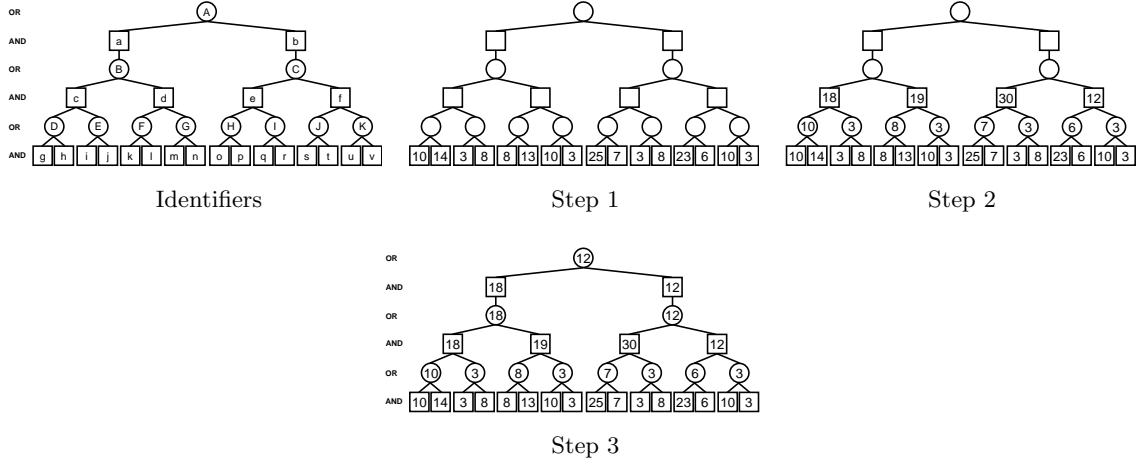

Figure 6: Simplified Trace of the Updates of the (Lower and Upper) Bounds

- **Step 2:** Agent $a_2$ updates the bounds of OR nodes $D$ to $K$ that it maintains to the bounds of the same OR nodes that leaf agents $a_3$ and $a_4$ maintain according to update equations 8 and 11, the bounds of AND nodes $c$ to $f$ to the sum of their delta costs and the bounds of their child OR nodes according to update equations 9 and 12 and the bounds of OR nodes $B$ and $C$ to the minimum of the bounds of their child AND nodes according to update equations 10 and 13. The bounds of OR nodes $B$ and $C$ are not shown in the figure since they are not (yet) maintained by agent $a_1$.

- **Step 3:** Agent $a_1$ updates the bounds of OR nodes $B$ and $C$ that it maintains to the bounds of the same OR nodes that agent $a_2$ maintains according to update equations 8 and 11, the bounds of AND nodes $a$ and $b$ to the sum of their delta costs and the bounds of their child OR nodes according to update equations 9 and 12 and the bounds of OR node $A$ to the minimum of the bounds of its child AND nodes according to update equations 10 and 13. Since the lower and upper bounds of a node are equal to its gamma cost, the lower and upper bounds of the root node are equal to its gamma cost, which in turn is equal to the minimal solution cost. The propagation terminates after three steps with minimal solution cost 12.

### 3.2.3 ADHERING TO MEMORY LIMITATIONS

Our description of BnB-ADOPT so far assumes no memory limitations. However, BnB-ADOPT is a memory-bounded DCOP search algorithm with memory requirements per agent that are linear in the number of agents. We now describe how BnB-ADOPT adheres to these memory limitations using techniques that were introduced for ADOPT but apply to BnB-ADOPT as well.

The simplified trace in Figure 6 assumes that every agent $a$ maintains its bounds for all values $d$, all child agents $c$ and all contexts $X^a$. The number of contexts can be exponential in the depth of the agent in the pseudo-tree. For our example DCOP problem, agent $a_3$ has four different contexts for the four different combinations of values of its ancestor agents $a_1$ and $a_2$. An agent cannot maintain an exponential number of bounds due to the memory limitations. Therefore, every agent maintains its bounds for only one context at any given time. This context is stored in the variable $X^a$ for agent $a$. The size of the context is at most linear in the number of agents. The number of bounds of an agent is now linear in the product of its domain cardinality and the number of its child agents. Thus, the memory requirements per agent are only linear in the number of agents if the domain cardinality and the magnitude of the bounds (and the other variables) are constant for each agent.

### 3.2.4 Performing Depth-First Search

Our description of BnB-ADOPT so far applies to ADOPT as well. However, BnB-ADOPT uses depth-first branch-and-bound search and ADOPT uses best-first search. We now describe how BnB-ADOPT implements depth-first search.

Agents of BnB-ADOPT send messages that are similar to that of ADOPT but processes them differently. They send messages of three different types, namely VALUE, COST and TERMINATE messages. At the start, every agent $a$ initializes its context $X^a$, uses update equations 6, 9, 10, 7, 12 and 13 to initialize its bounds and takes on its best value $d^a := \arg\min_{d \in Dom(a)} \{LB^a_{X^a}(d)\}$. It sends VALUE messages to all child agents and a COST message to its parent agent. It then repeatedly waits for incoming messages, processes them, possibly takes on a different value and again sends VALUE messages to all child agents and a COST message to its parent agent. A description of the three message types and how agents process them is as follows:

- **VALUE messages:** An agent $a$ with context $X^a$ and value $d^a$ sends VALUE messages to all child agents with the desired context $X^a \cup (a, d^a)$, which is its context augmented with its value. Leaf agents do not have child agents and thus do not send VALUE messages. VALUE messages thus propagate contexts down the pseudo-tree.

  When an agent receives a VALUE message, it checks whether its context is identical to the desired context in the VALUE message. If it is not, then the agent changes its context to the desired context in the VALUE message. In either case, it then executes the common program (see below).

- **COST messages:** An agent $a$ sends COST messages to its parent agent with its identity $a$, its context $X^a$ and its bounds $LB^a_{X^a}$ and $UB^a_{X^a}$. The root agent does not have a parent agent and thus does not send COST messages. COST messages thus propagate bounds up the pseudo-tree.

  When an agent receives a COST message, it checks whether its context and the context in the COST message are compatible. Two contexts are compatible if no agent takes on different values in the two contexts. If they are, then the agent uses update equations 8 to 13 with the bounds in the COST message to improve its bounds for the value in the message. In either case, it then executes the common program (see below).

- **TERMINATE messages:** When the termination condition $UB^r_{X^r} \leq LB^r_{X^r}$ is satisfied, the root agent $r$ sends TERMINATE messages (without parameters) to all child agents to inform them that the search is complete and then terminates. When an agent receives such a TERMINATE message, it sends TERMINATE messages to all child agents and terminates as well. Leaf agents do not have child agents and thus do not send TERMINATE messages. TERMINATE messages thus propagate down the pseudo-tree until all agents terminate.

The common program is as follows:

- **Context change:** If an agent $a$ changed its context $X^a$, it executes the following statements: It uses update equations 6, 9, 10, 7, 12 and 13 to initialize its bounds and takes on its best value $d^a := \arg\min_{d \in Dom(a)} \{LB^a_{X^a}(d)\}$. It then sends VALUE messages to all child agents and a COST message to its parent agent.

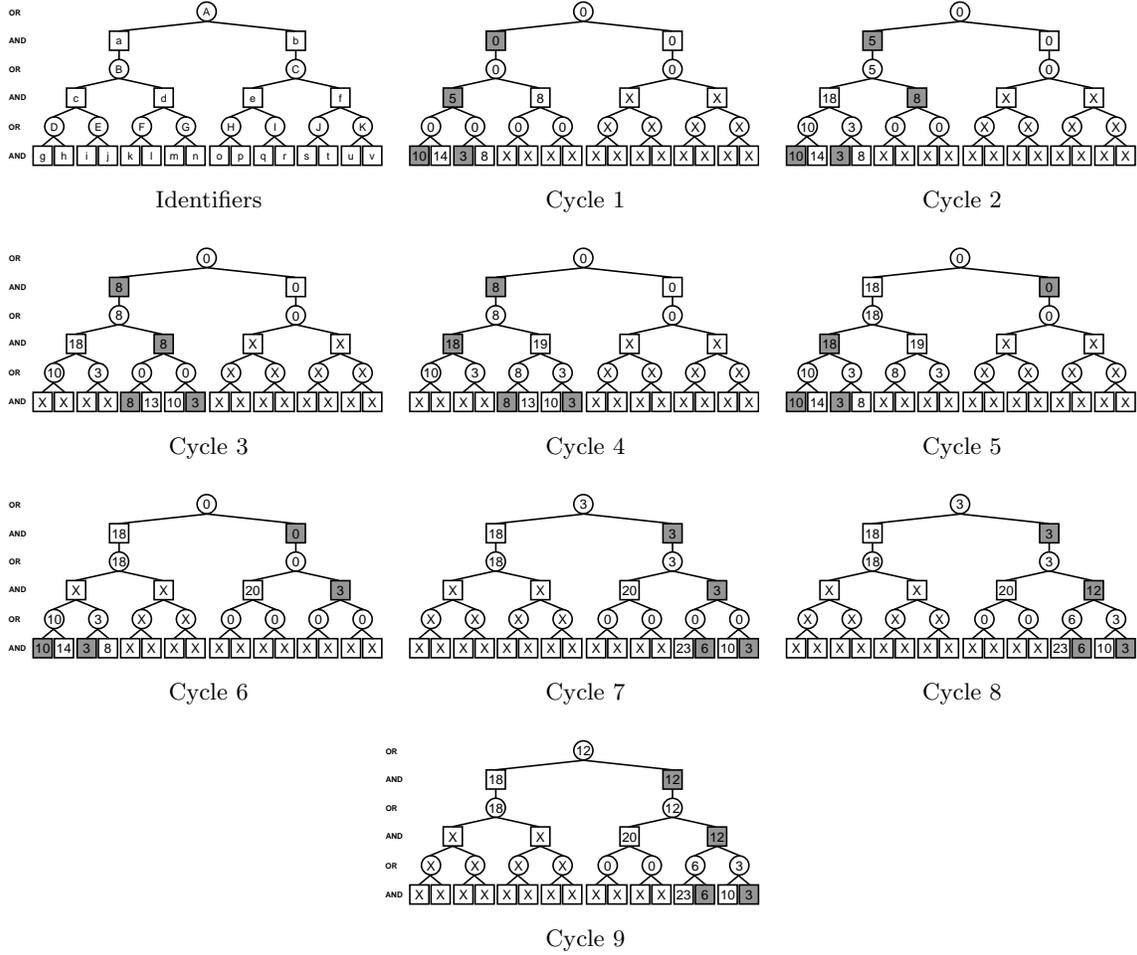

Figure 7: Trace of the Updates of the Lower Bounds

- **No context change:** If an agent $a$ did not change its context $X^a$, it executes the following statements: If $UB^a_{X^a} \leq LB^a_{X^a}(d^a)$ for its value $d^a$, then the context of the agent augmented with its value cannot be completed to a solution whose solution cost is smaller than the solution cost of the best solution found so far for its context $X^a (= UB^a_{X^a})$ and the agent thus takes on its best value $d^a := \arg\min_{d \in Dom(a)}\{LB^a_{X^a}(d)\}$. It then sends VALUE messages to all child agents and a COST message to its parent agent.

Assume that the context $X^a$ of an agent $a$ does not change. After a finite amount of time, $UB^a_{X^a} \leq LB^a_{X^a}(d^a)$ for its value $d^a$. The agent then takes on its best value and repeats the procedure. After a finite amount of time, $UB^a_{X^a} \leq LB^a_{X^a}(d)$ for all values $d$, which implies that $UB^a_{X^a} \leq LB^a_{X^a}$. The agent takes on every value $d$ at most once until $UB^a_{X^a} \leq LB^a_{X^a}$ since $LB^a_{X^a}(d)$ remains unchanged and $UB^a_{X^a}$ is monotonically non-increasing once the agent changes its value from $d$ to a different value, which prevents the agent from changing its value back to $d$ before $UB^a_{X^a} \leq LB^a_{X^a}$. BnB-ADOPT thus performs depth-first search. Then, after a finite amount of time, $UB^r_{X^r} \leq LB^r_{X^r}$ and the bound property $UB^r_{X^r} \geq LB^r_{X^r}$ together imply that $UB^r_{X^r} = \gamma^r_{X^r} = LB^r_{X^r}$ for the root agent $r$, and the DCOP problem is solved optimally.

Figures 7 and 8 show traces of the updates of the lower and upper bounds, respectively, for our example DCOP problem. BnB-ADOPT uses the zero heuristic values. The initial context of every agent assigns value 0 to all ancestor agents of the agent. We partition time into cycles. Agents maintain their bounds for only one context at any given time. Nodes in the figures are crossed out if their agent does not maintain their bounds. AND nodes are shaded if their partial solution is equal to the context of the agent of their parent OR node augmented with its value. For example, agents $a_1$, $a_3$ and $a_4$ take on value 0 in Cycle 2, and agent $a_2$ takes on value 1. The context of agent $a_1$ is $\{\}$, the context of agent $a_2$ is $\{(a_1, 0)\}$ and the contexts of agents $a_3$ and $a_4$ are $\{(a_1, 0), (a_2, 0)\}$. A description of the trace is as follows:

- **Cycle 1:** Root agent $a_1$ initializes its context $X^{a_1}$ to $\{\}$. It initializes the lower bounds of nodes $B$ $(= lb_{X^{a_1}}^{a_1,a_2}(0))$ and $C$ $(= lb_{X^{a_1}}^{a_1,a_2}(1))$ to 0 since it uses the zero heuristic values. It updates the lower bound of node $a$ $(= LB_{X^{a_1}}^{a_1}(0))$ to the sum of its delta cost $(= 0)$ and the lower bound of node $B$ $(= 0)$ according to the update equations. It updates the lower bound of node $b$ $(= LB_{X^{a_1}}^{a_1}(1))$ to the sum of its delta cost $(= 0)$ and the lower bound of node $C$ $(= 0)$ according to the update equations. It updates the lower bound of node $A$ $(= LB_{X^{a_1}}^{a_1})$ to the minimum of the lower bound of node $a$ $(= 0)$ and the lower bound of node $b$ $(= 0)$ according to the update equations. It initializes the upper bounds of nodes $B$ and $C$ to infinity. It updates the upper bounds of nodes $a$, $b$ and $A$ to infinity according to the update equations. It takes on its best value. It can take on either value 0 or value 1 since the lower bounds of nodes $a$ and $b$ are both 0. It takes on value 0 and sends a VALUE message to its child agent $a_2$.

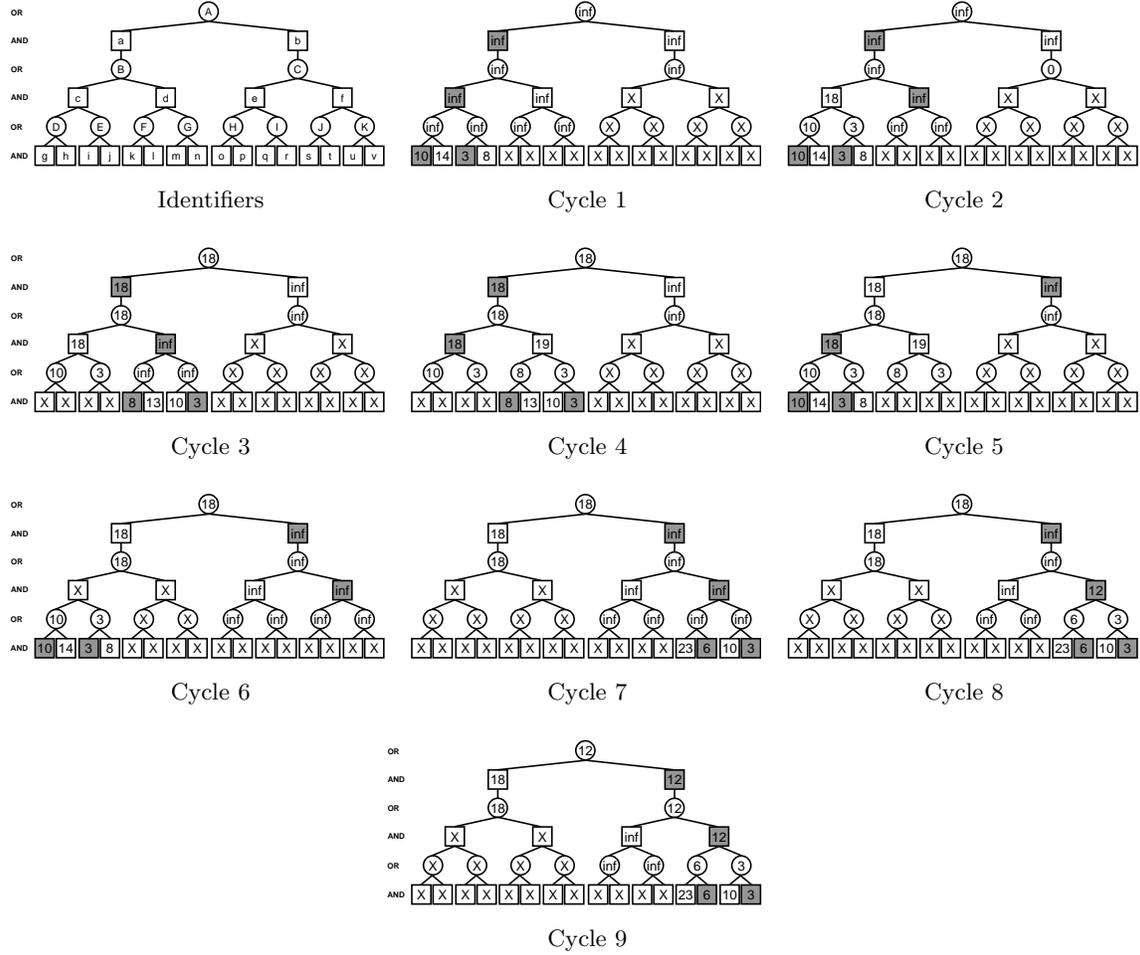

Figure 8: Trace of the Updates of the Upper Bounds

Agent $a_2$ initializes its context $X^{a_2}$ to $\{(a_1, 0)\}$. It initializes the lower bounds of nodes $D$, $E$, $F$ and $G$ to 0. It updates the lower bounds of nodes $c$, $d$ and $B$ to 5, 8 and 5, respectively. It initializes the upper bounds of nodes $D$, $E$, $F$ and $G$ to infinity. It updates the upper bounds of nodes $c$, $d$ and $B$ to infinity. The bounds of node $B$ that agent $a_2$ maintains are not shown in the figures. It takes on its best value 0, sends VALUE messages to its child agents $a_3$ and $a_4$ and sends a COST message to its parent agent $a_1$.

Leaf agent $a_3$ initializes its context $X^{a_3}$ to $\{(a_1, 0), (a_2, 0)\}$. It updates the lower bounds of nodes $g$ and $h$ to their delta costs 10 and 14, respectively, since leaf agents do not have child agents. It updates the lower bound of node $D$ to 10. It updates the upper bounds of nodes $g$ and $h$ to their delta costs 10 and 14, respectively, since leaf agents do not have child agents. It updates the upper bound of node $D$ to 10. The bounds of node $D$ that leaf agent $a_3$ maintains are not shown in the figures. It takes on its best value 0 and sends a COST message to its parent agent $a_2$.

Leaf agent $a_4$ initializes its context $X^{a_4}$ to $\{(a_1, 0), (a_2, 0)\}$. It updates the lower bounds of nodes $i$ and $j$ to their delta costs 3 and 8, respectively. It updates the lower bound of node $E$ to 3. It updates the upper bounds of nodes $i$ and $j$ to their delta costs 3 and 8, respectively. It updates the upper bound of node $E$ to 3. The bounds of node $E$ that leaf agent $a_4$ maintains are not shown in the figures. It takes on its best value 0 and sends a COST message to its parent agent $a_2$.

In summary, the following messages are sent during Cycle 1:

- message (VALUE, $\{(a_1, 0)\}$) from agent $a_1$ to agent $a_2$;
- message (VALUE, $\{(a_1, 0), (a_2, 0)\}$) from agent $a_2$ to agent $a_3$;
- message (VALUE, $\{(a_1, 0), (a_2, 0)\}$) from agent $a_2$ to agent $a_4$;
- message (COST, $a_2$, $\{(a_1, 0)\}$, 5, $\infty$) from agent $a_2$ to agent $a_1$;
- message (COST, $a_3$, $\{(a_1, 0), (a_2, 0)\}$, 10, 10) from agent $a_3$ to agent $a_2$; and
- message (COST, $a_4$, $\{(a_1, 0), (a_2, 0)\}$, 3, 3) from agent $a_4$ to agent $a_2$.

- **Cycle 2:** Root agent $a_1$ receives the COST message sent by its child agent $a_2$ in Cycle 1. Since the context of agent $a_1$ ($= \{\}$) is compatible with the context in the message ($= \{(a_1, 0)\}$), it improves its bounds. It updates the bounds of node $B$ to the bounds in the message ($= 5$ and infinity, respectively). It updates the bounds of nodes $a$, $b$ and $A$. It does not change its value since the lower bound of node $a$ ($= LB^{a_1}_{X^{a_1}}(d^{a_1}) = 5$ for its value $d^{a_1} = 0$) is still smaller than the upper bound of node $A$ ($= UB^{a_1}_{X^{a_1}} = \infty$). It sends a VALUE message to its child agent $a_2$.

Agent $a_2$ receives the VALUE message sent by its parent agent $a_1$ in Cycle 1. Its context ($= \{(a_1, 0)\}$) remains unchanged since it is the same as the desired context in the message ($= \{(a_1, 0)\}$). Agent $a_2$ also receives the COST messages sent by its child agents $a_3$ and $a_4$ in Cycle 1. Since the context of agent $a_2$ ($= \{(a_1, 0)\}$) is compatible with the contexts in the messages ($= \{(a_1, 0), (a_2, 0)\}$), it improves its bounds. It updates the bounds of node $D$ to the bounds in the first message ($= 10$ and 10, respectively) and the bounds of node $E$ to the bounds in the second message ($= 3$ and 3, respectively). It updates the bounds of nodes $c$, $d$ and $B$. It changes its value since the lower bound of node $c$ ($= LB^{a_2}_{X^{a_2}}(d^{a_2}) = 18$ for its value $d^{a_2} = 0$) is no longer smaller than the upper bound of node $B$ ($= UB^{a_2}_{X^{a_2}} = 18$). It takes on its best value 1, sends VALUE messages to its child agents $a_3$ and $a_4$ and sends a COST message to its parent agent $a_1$.

Leaf agents $a_3$ and $a_4$ receive the VALUE messages sent by their parent agent $a_2$ in Cycle 1. Their contexts $(= \{(a_1, 0), (a_2, 0)\})$ remain unchanged since they are the same as the desired context in the message $(= \{(a_1, 0), (a_2, 0)\})$. They send the same COST messages as before to their parent agent $a_2$.

In summary, the messages sent during Cycle 2 are identical to the ones sent during Cycle 1, except for the messages sent by agent $a_2$, which are as follows:

- message (VALUE, $\{(a_1, 0), (a_2, 1)\}$) from agent $a_2$ to agent $a_3$;

- message (VALUE, $\{(a_1, 0), (a_2, 1)\}$) from agent $a_2$ to agent $a_4$; and

- message (COST, $a_2$, $\{(a_1, 0)\}$, 8, 18) from agent $a_2$ to agent $a_1$.

The VALUE messages are different because agent $a_2$ changed its value from 0 to 1. The COST message is different because agent $a_2$ changed its bounds.

- **Cycles 3-9:** The messages sent during Cycle 3 are identical to the ones sent during Cycle 2, except for the messages sent by agents $a_3$ and $a_4$, which are as follows:

  - message (COST, $a_3$, $\{(a_1, 0), (a_2, 1)\}$, 8, 8) from agent $a_3$ to agent $a_2$; and

  - message (COST, $a_4$, $\{(a_1, 0), (a_2, 1)\}$, 3, 3) from agent $a_4$ to agent $a_2$.

  The COST messages are different because agents $a_3$ and $a_4$ changed their contexts. The termination condition holds after a finite amount of time when the upper bound of node $A$ $(= UB^{a_1}_{X^{a_1}} = 12)$ is no larger than the lower bound of node $A$ $(= LB^{a_1}_{X^{a_1}} = 12)$. Root agent $a_1$ sends TERMINATE messages to all child agents, and the TERMINATE messages propagate down the pseudo-tree until all agents terminate. BnB-ADOPT terminates after nine cycles with minimal solution cost 12.

### 3.2.5 Performing Branch-and-Bound

We now refine our description of BnB-ADOPT by explaining how the agents implement branch-and-bound search to make BnB-ADOPT faster. Every agent $a$ of BnB-ADOPT now also maintains the variable threshold $TH^a_{X^a}$, which it initializes to infinity. The threshold of the root agent always remains infinity. Every other agent uses its threshold for pruning, meaning that it can change its value earlier than previously.

- **First change:** If an agent $a$ did not change its context $X^a$, it previously executed the following statements: If $UB^a_{X^a} \leq LB^a_{X^a}(d^a)$ for its value $d^a$, then the agent took on its best value. It then sent VALUE messages to all child agents and a COST message to its parent agent. Now, if $TH^a_{X^a} \leq LB^a_{X^a}(d^a)$, then the agent also takes on its best value. Thus, if $\min\{TH^a_{X^a}, UB^a_{X^a}\} \leq LB^a_{X^a}(d^a)$, then the agent takes on its best value and thus potentially changes its value, which is earlier than previously. $\min\{TH^a_{X^a}, UB^a_{X^a}\}$ is the pruning quantity.

- **Second change:** An agent $a$ with context $X^a$ and value $d^a$ sends VALUE messages to all its child agents, which previously contained only the desired context $X^a \cup (a, d^a)$. VALUE messages now also contain the desired threshold $\min\{TH^a_{X^a}, UB^a_{X^a}\} - \delta^a_{X^a}(d^a) - \sum_{c' \in C(a) \backslash c} lb^{a,c'}_{X^a}(d^a)$ for the child agent $c$. When agent $c$ receives a VALUE message, it sets its threshold to the desired threshold and then proceeds as described earlier. The desired threshold is set such that the lower bound $LB^a_{X^a}(d^a)$ of agent $a$ for its value $d^a$ reaches its pruning quantity (and agent $a$ thus potentially changes its value) when the lower bound $LB^c_{X^c}$ of agent $c$ reaches the desired threshold. This property can be verified as follows:

$$LB_{X^c}^c \geq \min\{TH_{X^a}^a, UB_{X^a}^a\} - \delta_{X^a}^a(d^a) - \sum_{c' \in C(a) \setminus c} lb_{X^a}^{a,c'}(d^a) \tag{14}$$

$$lb_{X^a}^{a,c}(d^a) \geq \min\{TH_{X^a}^a, UB_{X^a}^a\} - \delta_{X^a}^a(d^a) - \sum_{c' \in C(a) \setminus c} lb_{X^a}^{a,c'}(d^a) \tag{15}$$

$$-\min\{TH_{X^a}^a, UB_{X^a}^a\} \geq -\delta_{X^a}^a(d^a) - lb_{X^a}^{a,c}(d^a) - \sum_{c' \in C(a) \setminus c} lb_{X^a}^{a,c'}(d^a) \tag{16}$$

$$\min\{TH_{X^a}^a, UB_{X^a}^a\} \leq \delta_{X^a}^a(d^a) + lb_{X^a}^{a,c}(d^a) + \sum_{c' \in C(a) \setminus c} lb_{X^a}^{a,c'}(d^a) \tag{17}$$

$$\min\{TH_{X^a}^a, UB_{X^a}^a\} \leq \delta_{X^a}^a(d^a) + \sum_{c' \in C(a)} lb_{X^a}^{a,c'}(d^a) \tag{18}$$

$$\min\{TH_{X^a}^a, UB_{X^a}^a\} \leq LB_{X^a}^a(d^a) \tag{19}$$

### 3.2.6 Further Enhancements

We continue to refine our description of BnB-ADOPT by explaining a number of additional enhancements, which were introduced for ADOPT.

- **Reduced contexts:** The agents now use reduced contexts, which are subsets of the contexts described previously. The reduced context $X_1^a$ of agent $a$ contains the values of all ancestor agents $p \in SCP(a)$, while the context $X_2^a$ described previously contains the values of all ancestor agents $p \in P(a)$. The agents can use reduced contexts since $\gamma_{X_1^a}^a = \gamma_{X_2^a}^a$ and $\gamma_{X_1^a}^a(d) = \gamma_{X_2^a}^a(d)$ for all values $d$. Agents now use reduced contexts because they need to change their contexts and thus initialize their bounds less often when they receive VALUE messages since their contexts are then more often identical to the desired contexts in the VALUE messages. For our example DCOP problem, the reduced context of agent $a_4$ contains the values of only agent $a_2$ rather than the values of agents $a_1$ and $a_2$. Therefore, the following pairs of nodes in the search tree are actually the same node: nodes $i$ and $q$, nodes $j$ and $r$, nodes $m$ and $u$, and nodes $n$ and $v$.

- **VALUE and COST messages:** An agent sends VALUE messages to all child agents, which previously contained the desired context and the desired threshold. The desired context is the context of the agent augmented with its value. When an agent receives a VALUE message, it previously checked whether its context is identical to the desired context in the VALUE message. If it was not, then the agent changed its context to the desired context in the VALUE message. Agents now update their contexts differently to reduce the size of the VALUE messages. An agent sends VALUE messages to all child and pseudo-child agents with its identity, value and desired threshold, which is infinity for its pseudo-child agents. When an agent receives a VALUE message, it sets its threshold to the desired threshold if the message is from its parent agent. It also checks whether the value of the ancestor agent in the VALUE message is more recent than the value of the ancestor agent in its context. If it is, then the agent changes the value of the ancestor agent in its context to the value of the ancestor agent in the VALUE message. However, the context of an agent does not only contain the values of its parent and pseudo-parent agents but also the values of its ancestor agents that are the parent or pseudo-parent agents of one (or more) of its descendant agents, and ancestor agents that are not constrained with the agent cannot send VALUE messages to the agent. However, they send VALUE messages to their pseudo-child agents, at least one of which is a descendant agent of the agent, and the information then propagates up the pseudo-tree with COST messages until it reaches the agent. When an agent receives a COST message, it now checks whether the value of an ancestor agent in the context of the COST message is more recent than the value of the ancestor agent in its context. If it is, then the agent changes the value of the ancestor agent in its context to the value of the ancestor agent in the context of the COST message. Our example DCOP problem is too simple to allow us to illustrate the propagation of the information up the pseudo-tree. However, imagine that a new agent $a_5$ is a child agent of agent $a_4$ and is constrained with agents $a_1$ and $a_4$. The context of agent $a_4$ then contains the value of agent $a_1$ but agent $a_1$ cannot send VALUE messages to agent $a_4$. However, agent $a_1$ sends VALUE messages to agent $a_5$. Agent $a_5$ changes the value of agent $a_1$ in its context and sends COST messages with its context to agent $a_4$, which then changes the value of agent $a_1$ in its context as well.

The agents now need to determine whether the value of an agent in VALUE messages or in the contexts of COST messages is more recent than the value of the agent in their contexts. Every agent $a$ therefore now also maintains a counter $ID^a$ and increments it whenever it changes its value. Therefore, a larger ID indicates a more recent value. The values of agents in contexts are now labeled with their IDs, and VALUE messages contain the identity of the sending agent, its value, its ID and the desired threshold.

- **Bounds:** Whenever an agent changes its context $X^a$, it previously initialized its bounds and took on its best value. The (reduced) context of a child agent of an agent can now be a strict subset of the (reduced) context of the agent since the parent or some pseudo-parent agents of the agent might not be (parent or) pseudo-parent agents of the child agent or its descendant agents. If the context of child agent $c$ does not contain the values of any agents whose values changed in the context of agent $a$, then agent $a$ does not initialize its lower bounds $lb_{X^a}^{a,c}(d)$ and upper bounds $ub_{X^a}^{a,c}(d)$ for agent $c$ and all values $d$ before it takes on its best value. Agents use this optimization because they need to initialize their bounds less often this way. For our example DCOP problem, if agent $a_2$ changes its context from $\{(a_1, 0)\}$ to $\{(a_1, 1)\}$ (where the IDs are omitted for simplicity), then it does not initialize its lower bounds $lb_{X^{a_2}}^{a_2,a_4}(d)$ and upper bounds $ub_{X^{a_2}}^{a_2,a_4}(d)$ for child agent $a_4$ and all values $d$ since the context of agent $a_4$ does not contain the value of agent $a_1$.

Additionally, if an agent $a$ changes its context due to a COST message from its child agent $c$ and its new context $X^a$ is compatible with the context in the COST message, then agent $a$ can set its lower bound $lb_{X^a}^{a,c}(d)$ and upper bound $ub_{X^a}^{a,c}(d)$ for agent $c$ and value $d$ of agent $a$ in the COST message to the bounds in the COST message before it takes on its best value. Agents use this optimization because the bounds in the COST message are more informed than the initialized bounds. Our example DCOP problem is too simple to allow us to illustrate this optimization. However, imagine again that a new agent $a_5$ is a child agent of agent $a_4$ and is constrained with agents $a_1$ and $a_4$. Assume that the context of agent $a_4$ is $\{(a_1, 0), (a_2, 0)\}$ (where the IDs are again omitted for simplicity) and it receives a COST message from agent $a_5$ with context $\{(a_1, 1), (a_4, 0)\}$. Agent $a_4$ then changes its context to $\{(a_1, 1), (a_2, 0)\}$, sets its lower bound $lb_{\{(a_1,1),(a_2,0)\}}^{a_4,a_5}(0)$ and its upper bound $ub_{\{(a_1,1),(a_2,0)\}}^{a_4,a_5}(0)$ to the bounds in the COST message and initializes all other bounds before it takes on its best value.

### 3.2.7 Pseudocode

Figure 9 shows the BnB-ADOPT pseudocode of every agent. The pseudocode does not index variables with the context since this context is implicitly given by the variable $X^a$. It uses the predicate Compatible$(X, X') = \neg \exists_{(a,d,ID)\in X,(a',d',ID')\in X'}(a = a' \wedge d \neq d')$ that determines if two contexts $X$ and $X'$ are compatible, that is, if no agent takes on two different values in the two contexts [Lines 35, 44, 46, 48 and 51]. The pseudocode also uses the procedure PriorityMerge$(X, X')$ that executes $X' := \{(a', d', ID') \in X' \mid \neg \exists_{(a,d,ID)\in X}(a = a')\} \cup \{(a', d', ID') \in X' \mid \exists_{(a,d,ID)\in X}(a =$

**procedure Start**()
[01]  $X^a := \{(p, \textbf{ValInit}(p), 0) \mid p \in SCP(a)\}$;
[02]  $ID^a := 0$;
[03]  **forall** $c \in C(a), d \in Dom(a)$
[04]    **InitChild**($c, d$);
[05]  **InitSelf**();
[06]  **Backtrack**();
[07]  **loop forever**
[08]    **if**(message queue is not empty)
[09]      **while**(message queue is not empty)
[10]        pop $msg$ off message queue;
[11]        **When Received**($msg$);
[12]      **Backtrack**();

**procedure InitChild**($c, d$)
[13]  $lb^{a,c}(d) := h^{a,c}(d)$;
[14]  $ub^{a,c}(d) := \infty$;

**procedure InitSelf**()
[15]  $d^a := \arg\min_{d \in Dom(a)}\{\delta^a(d) + \sum_{c \in C(a)} lb^{a,c}(d)\}$;
[16]  $ID^a := ID^a + 1$;
[17]  $TH^a := \infty$;

**procedure Backtrack**()
[18]  **forall** $d \in Dom(a)$
[19]    $LB^a(d) := \delta^a(d) + \sum_{c \in C(a)} lb^{a,c}(d)$;
[20]    $UB^a(d) := \delta^a(d) + \sum_{c \in C(a)} ub^{a,c}(d)$;
[21]  $LB^a := \min_{d \in Dom(a)}\{LB^a(d)\}$;
[22]  $UB^a := \min_{d \in Dom(a)}\{UB^a(d)\}$;
[23]  **if**($LB^a(d^a) \geq \min\{TH^a, UB^a\}$)
[24]    $d^a := \arg\min_{d \in Dom(a)}\{LB^a(d)\}$ (choose the previous $d^a$ if possible);
[25]    **if** a new $d^a$ has been chosen
[26]      $ID^a := ID^a + 1$;
[27]  **if**(($a$ is root and $UB^a \leq LB^a$) or termination message received)
[28]    **Send**(TERMINATE) to each $c \in C(a)$;
[29]    terminate execution;
[30]  **Send**(VALUE, $a, d^a, ID^a, \min\{TH^a, UB^a\} - \delta^a(d^a) - \sum_{c' \in C(a)\setminus c} lb^{a,c'}(d^a)$) to each $c \in C(a)$;
[31]  **Send**(VALUE, $a, d^a, ID^a, \infty$) to each $c \in CD(a) \setminus C(a)$;
[32]  **Send**(COST, $a, X^a, LB^a, UB^a$) to $pa(a)$ if $a$ is not root;

**procedure When Received**(VALUE, $p, d^p, ID^p, TH^p$)
[33]  $X' := X^a$;
[34]  **PriorityMerge**($(p, d^p, ID^p), X^a$);
[35]  **if**(!**Compatible**($X', X^a$))
[36]    **forall** $c \in C(a), d \in Dom(a)$
[37]      **if**($p \in SCP(c)$)
[38]        **InitChild**($c, d$);
[39]    **InitSelf**();
[40]  **if**($p = pa(a)$)
[41]    $TH^a := TH^p$;

**procedure When Received**(COST, $c, X^c, LB^c, UB^c$)
[42]  $X' := X^a$;
[43]  **PriorityMerge**($X^c, X^a$);
[44]  **if**(!**Compatible**($X', X^a$))
[45]    **forall** $c \in C(a), d \in Dom(a)$
[46]      **if**(!**Compatible**($\{(p, d^p, ID^p) \in X' \mid p \in SCP(c)\}, X^a$))
[47]        **InitChild**($c, d$);
[48]  **if**(**Compatible**($X^c, X^a$))
[49]    $lb^{a,c}(d) := \max\{lb^{a,c}(d), LB^c\}$ for the unique $(a', d, ID) \in X^c$ with $a' = a$;
[50]    $ub^{a,c}(d) := \min\{ub^{a,c}(d), UB^c\}$ for the unique $(a', d, ID) \in X^c$ with $a' = a$;
[51]  **if**(!**Compatible**($X', X^a$))
[52]    **InitSelf**();

**procedure When Received**(TERMINATE)
[53]  record termination message received;

Figure 9: Pseudocode of BnB-ADOPT

$a' \wedge ID \leq ID')\} \cup \{(a, d, ID) \in X \mid \exists_{(a', d', ID') \in X'}(a = a' \wedge ID > ID')\}$ and thus replaces the values of agents in context $X'$ with more recent values, if available, of the same agents in context $X$ [Lines 34 and 43].

The code is identical for every agent except that the variable $a$ is a "self" variable that points to the agent itself. At the start, BnB-ADOPT calls Start() for every agent. When an agent $a$ receives a VALUE message from an ancestor agent, then the "When Received" handler for VALUE messages is called with $p$ being the ancestor agent, $d^p$ being the value of the ancestor agent, $ID^p$ being the ID of the ancestor agent and $TH^p$ being the desired threshold for agent $a$ if the ancestor agent is its parent agent (and infinity otherwise) [Line 11]. When agent $a$ receives a COST message from a child agent, then the "When Received" handler for COST messages is called with $c$ being the child agent, $X^c$ being the context of the child agent, $LB^c$ being the lower bound $LB^c_{X^c}$ of the child agent and $UB^c$ being the upper bound $UB^c_{X^c}$ of the child agent [Line 11]. Finally, when agent $a$ receives a TERMINATE message from its parent agent, then the "When Received" handler for TERMINATE messages is called without any arguments [Line 11].

BnB-ADOPT uses the same message passing and communication framework as ADOPT and has the same memory requirements. It uses similar VALUE, COST and TERMINATE messages, a similar strategy to update the context of an agent based on VALUE messages from its ancestor agents and COST messages from its child agents, the same semantics for the bounds and the same update equations to update these bounds. BnB-ADOPT and ADOPT both use thresholds but BnB-ADOPT uses the thresholds for pruning while ADOPT uses them to reconstruct partial solutions that were purged from memory. Thus, BnB-ADOPT uses a different threshold initialization [Line 17], different desired threshold calculation [Line 30] and different termination condition [Line 27]. BnB-ADOPT also differs from ADOPT in that it maintains IDs that agents use to indicate the recency of their values and labels the values of agents in contexts with their IDs.

### 3.2.8 Trace

Figures 10 and 11 show traces of the updates of the lower and upper bounds, respectively, for our example DCOP problem, and Table 2 shows a trace of the update of all variables. BnB-ADOPT uses the heuristic values $h^{a_1,a_2}_{X^{a_1}}(0) := 3$, $h^{a_1,a_2}_{X^{a_1}}(1) := 6$, $h^{a_2,a_3}_{X^{a_2}}(0) := 2$, $h^{a_2,a_3}_{X^{a_2}}(1) := 2$, $h^{a_2,a_4}_{X^{a_2}}(0) := 2$ and $h^{a_2,a_4}_{X^{a_2}}(1) := 2$ for all contexts $X^{a_1}$ and $X^{a_2}$. These heuristic values were chosen by hand. Every agent assigns the value of all its ancestor agents in its initial context to 0. We partition time into cycles as in Figures 7 and 8 and continue to use the conventions made in the context of those figures.

- **Cycle 1:** Root agent $a_1$ initializes its context $X^{a_1}$ to {} [Line 1]. It initializes the lower bounds of nodes $B$ ($= lb^{a_1,a_2}_{X^{a_1}}(0)$) and $C$ ($= lb^{a_1,a_2}_{X^{a_1}}(1)$) to their heuristic values 3 and 6, respectively [Line 13]. It updates the lower bound of node $a$ ($= LB^{a_1}_{X^{a_1}}(0)$) to the sum of its delta cost ($= 0$) and the lower bound of node $B$ ($= 3$) according to the update equations [Line 19]. It updates the lower bound of node $b$ ($= LB^{a_1}_{X^{a_1}}(1)$) to the sum of its delta cost ($= 0$) and the lower bound of node $C$ ($= 6$) according to the update equations [Line 19]. It updates the lower bound of node $A$ ($= LB^{a_1}_{X^{a_1}}$) to the minimum of the lower bound of node $a$ ($= 3$) and the lower bound of node $b$ ($= 6$) according to the update equations [Line 21]. It initializes the upper bounds of nodes $B$ and $C$ to infinity [Line 14]. It updates the upper bounds of nodes $a$, $b$ and $A$ to infinity according to the update equations [Lines 20 and 22]. It takes on its best value 0 since the lower bound of node $a$ ($= 3$) is smaller than the lower bound of node $b$ ($= 6$) [Line 15], initializes its ID $ID^{a_1}$ to 1 [Lines 2 and 16], initializes its threshold $TH^{a_1}$ to infinity [Line 17] and sends VALUE messages to its child agent $a_2$ and pseudo-child agent $a_3$ [Lines 30 and 31].

  Agent $a_2$ initializes its context $X^{a_2}$ to $\{(a_1, 0, 0)\}$ [Line 1]. It initializes the lower bounds of nodes $D$, $E$, $F$ and $G$ to their heuristic value 2 [Line 13]. It updates the lower bounds of nodes $c$, $d$ and $B$ to 9, 12 and 9, respectively [Lines 19 and 21]. It initializes the upper bounds of nodes $D$, $E$, $F$ and $G$ to infinity [Line 14]. It updates the upper bounds of nodes $c$, $d$ and $B$ to infinity [Lines 20 and 22]. The bounds of node $B$ that agent $a_2$ maintains are not shown in the figure. It takes on its best value 0 [Line 15], initializes its ID to 1 [Lines 2 and 16], initializes its threshold to infinity [Line 17] and sends VALUE messages to its child agents $a_3$ and $a_4$ and a COST message to its parent agent $a_1$ [Lines 30-32].

| Cycle | 1 | 2 | 3 | 4 | 5 | 6 | 7 | 8 | 9 |
|---|---|---|---|---|---|---|---|---|---|
| $X^{a_1}$ | | | | | | | | | |
| $d^{a_1}$ | 0 | 0 | 0 | 0 | 1 | 1 | 1 | 1 | 1 |
| $ID^{a_1}$ | 1 | 1 | 1 | 1 | 2 | 2 | 2 | 2 | 2 |
| $TH^{a_1}$ | ∞ | ∞ | ∞ | ∞ | ∞ | ∞ | ∞ | ∞ | ∞ |
| $LB^{a_1}(0)$ | 3 | 9 | 12 | 12 | 18 | 18 | 18 | 18 | 18 |
| $LB^{a_1}(1)$ | 6 | 6 | 6 | 6 | 6 | 6 | 8 | 8 | 12 |
| $LB^{a_1}$ | 3 | 6 | 6 | 6 | 6 | 6 | 8 | 8 | 12 |
| $UB^{a_1}(0)$ | ∞ | ∞ | 18 | 18 | 18 | 18 | 18 | 18 | 18 |
| $UB^{a_1}(1)$ | ∞ | ∞ | ∞ | ∞ | ∞ | ∞ | ∞ | 30 | 12 |
| $UB^{a_1}$ | ∞ | ∞ | 18 | 18 | 18 | 18 | 18 | 18 | 12 |
| $lb^{a_1,a_2}(0)$ | 3 | 9 | 12 | 12 | 18 | 18 | 18 | 18 | 18 |
| $lb^{a_1,a_2}(1)$ | 6 | 6 | 6 | 6 | 6 | 6 | 8 | 8 | 12 |
| $ub^{a_1,a_2}(0)$ | ∞ | ∞ | 18 | 18 | 18 | 18 | 18 | 18 | 18 |
| $ub^{a_1,a_2}(1)$ | ∞ | ∞ | ∞ | ∞ | ∞ | ∞ | ∞ | 30 | 12 |
| $X^{a_2}$ | $(a_1,0,0)$ | $(a_1,0,1)$ | $(a_1,0,1)$ | $(a_1,0,1)$ | $(a_1,0,1)$ | $(a_1,1,2)$ | $(a_1,1,2)$ | $(a_1,1,2)$ | $(a_1,1,2)$ |
| $d^{a_2}$ | 0 | 1 | 1 | 0 | 0 | 1 | 1 | 1 | 1 |
| $ID^{a_2}$ | 1 | 2 | 2 | 3 | 3 | 4 | 4 | 4 | 4 |
| $TH^{a_2}$ | ∞ | ∞ | ∞ | 18 | 18 | 18 | 18 | 18 | 18 |
| $LB^{a_2}(0)$ | 9 | 18 | 18 | 18 | 18 | 25 | 30 | 30 | 30 |
| $LB^{a_2}(1)$ | 12 | 12 | 12 | 19 | 19 | 8 | 8 | 12 | 12 |
| $LB^{a_2}$ | 9 | 12 | 12 | 18 | 18 | 8 | 8 | 12 | 12 |
| $UB^{a_2}(0)$ | ∞ | 18 | 18 | 18 | 18 | ∞ | 30 | 30 | 30 |
| $UB^{a_2}(1)$ | ∞ | ∞ | ∞ | 19 | 19 | ∞ | ∞ | 12 | 12 |
| $UB^{a_2}$ | ∞ | 18 | 18 | 18 | 18 | ∞ | 30 | 12 | 12 |
| $lb^{a_2,a_3}(0)$ | 2 | 10 | 10 | 10 | 10 | 2 | 7 | 7 | 7 |
| $lb^{a_2,a_3}(1)$ | 2 | 2 | 2 | 8 | 8 | 2 | 2 | 6 | 6 |
| $ub^{a_2,a_3}(0)$ | ∞ | 10 | 10 | 10 | 10 | ∞ | 7 | 7 | 7 |
| $ub^{a_2,a_3}(1)$ | ∞ | ∞ | ∞ | 8 | 8 | ∞ | ∞ | 6 | 6 |
| $lb^{a_2,a_4}(0)$ | 2 | 3 | 3 | 3 | 3 | 3 | 3 | 3 | 3 |
| $lb^{a_2,a_4}(1)$ | 2 | 2 | 2 | 3 | 3 | 3 | 3 | 3 | 3 |
| $ub^{a_2,a_4}(0)$ | ∞ | 3 | 3 | 3 | 3 | 3 | 3 | 3 | 3 |
| $ub^{a_2,a_4}(1)$ | ∞ | ∞ | ∞ | 3 | 3 | 3 | 3 | 3 | 3 |
| $X^{a_3}$ | $(a_1,0,0)$ | $(a_1,0,1)$ | $(a_1,0,1)$ | $(a_1,0,1)$ | $(a_1,0,1)$ | $(a_1,1,2)$ | $(a_1,1,2)$ | $(a_1,1,2)$ | $(a_1,1,2)$ |
| | $(a_2,0,0)$ | $(a_2,0,1)$ | $(a_2,1,2)$ | $(a_2,1,2)$ | $(a_2,0,3)$ | $(a_2,0,3)$ | $(a_2,1,4)$ | $(a_2,1,4)$ | $(a_2,1,4)$ |
| $d^{a_3}$ | 0 | 0 | 0 | 0 | 0 | 1 | 1 | 1 | 1 |
| $ID^{a_3}$ | 1 | 1 | 2 | 2 | 3 | 4 | 5 | 5 | 5 |
| $TH^{a_3}$ | ∞ | ∞ | 8 | 8 | 10 | 10 | 12 | 12 | 6 |
| $LB^{a_3}(0)$ | 10 | 10 | 8 | 8 | 10 | 25 | 23 | 23 | 23 |
| $LB^{a_3}(1)$ | 14 | 14 | 13 | 13 | 14 | 7 | 6 | 6 | 6 |
| $LB^{a_3}$ | 10 | 10 | 8 | 8 | 10 | 7 | 6 | 6 | 6 |
| $UB^{a_3}(0)$ | 10 | 10 | 8 | 8 | 10 | 25 | 23 | 23 | 23 |
| $UB^{a_3}(1)$ | 14 | 14 | 13 | 13 | 14 | 7 | 6 | 6 | 6 |
| $UB^{a_3}$ | 10 | 10 | 8 | 8 | 10 | 7 | 6 | 6 | 6 |
| $X^{a_4}$ | $(a_2,0,0)$ | $(a_2,0,1)$ | $(a_2,1,2)$ | $(a_2,1,2)$ | $(a_2,0,3)$ | $(a_2,0,3)$ | $(a_2,1,4)$ | $(a_2,1,4)$ | $(a_2,1,4)$ |
| $d^{a_4}$ | 0 | 0 | 1 | 1 | 0 | 0 | 1 | 1 | 1 |
| $ID^{a_4}$ | 1 | 1 | 2 | 2 | 3 | 3 | 4 | 4 | 4 |
| $TH^{a_4}$ | ∞ | ∞ | 8 | 8 | 3 | 3 | 13 | 13 | 3 |
| $LB^{a_4}(0)$ | 3 | 3 | 10 | 10 | 3 | 3 | 10 | 10 | 10 |
| $LB^{a_4}(1)$ | 8 | 8 | 3 | 3 | 8 | 8 | 3 | 3 | 3 |
| $LB^{a_4}$ | 3 | 3 | 3 | 3 | 3 | 3 | 3 | 3 | 3 |
| $UB^{a_4}(0)$ | 3 | 3 | 10 | 10 | 3 | 3 | 10 | 10 | 10 |
| $UB^{a_4}(1)$ | 8 | 8 | 3 | 3 | 8 | 8 | 3 | 3 | 3 |
| $UB^{a_4}$ | 3 | 3 | 3 | 3 | 3 | 3 | 3 | 3 | 3 |

Table 2: Trace of the Update of all Variables of BnB-ADOPT

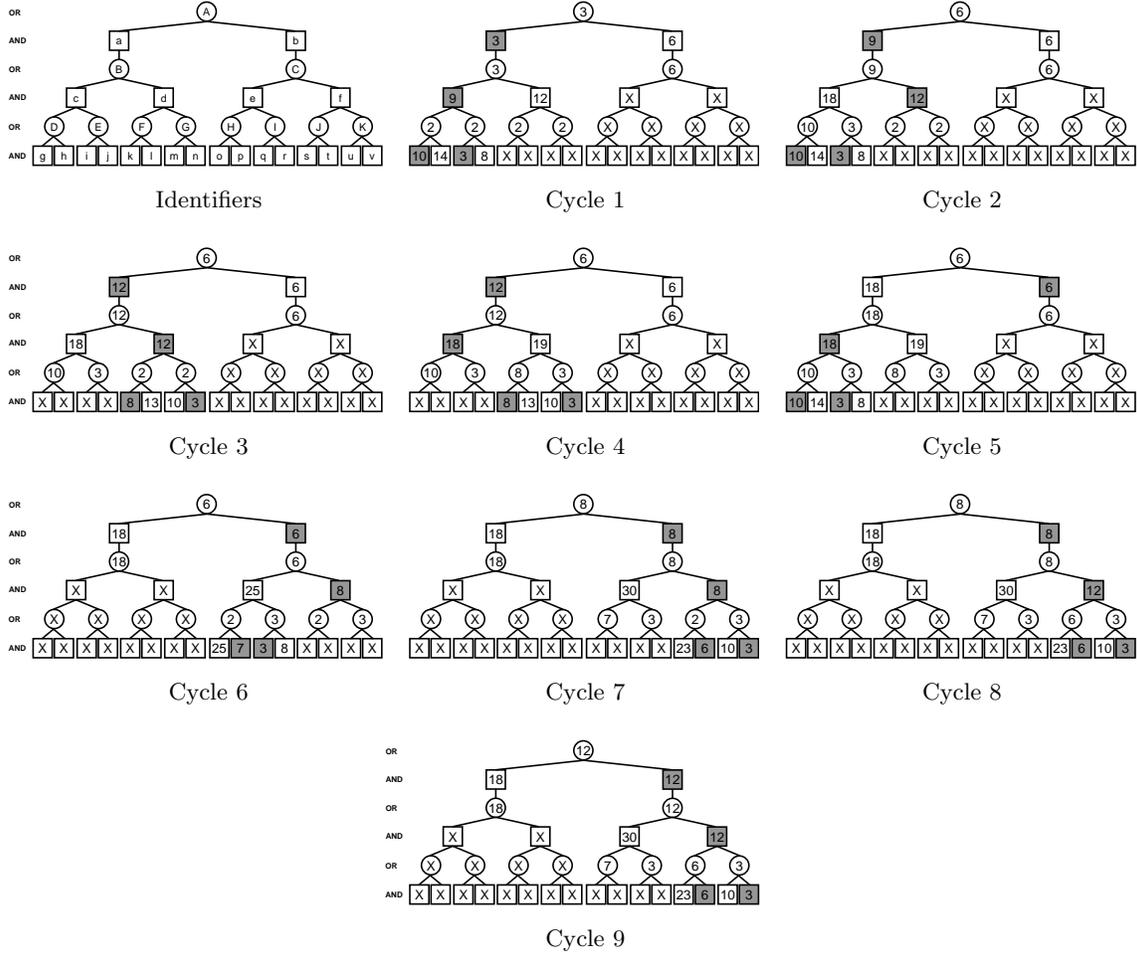

Figure 10: Trace of the Update of the Lower Bounds of BnB-ADOPT

Leaf agent $a_3$ initializes its context $X^{a_3}$ to $\{(a_1, 0, 0), (a_2, 0, 0)\}$ [Line 1]. It updates the lower bounds of nodes $g$ and $h$ to their delta costs 10 and 14, respectively, since leaf agents do not have child agents [Line 19]. It updates the lower bound of node $D$ to 10 [Line 21]. It updates the upper bounds of nodes $g$ and $h$ to their delta costs 10 and 14, respectively, since leaf agents do not have child agents [Line 20]. It updates the upper bound of node $D$ to 10 [Line 22]. The bounds of node $D$ that leaf agent $a_3$ maintains are not shown in the figure. It takes on its best value 0 [Line 15], initializes its ID to 1 [Lines 2 and 16], initializes its threshold to infinity [Line 17] and sends a COST message to its parent agent $a_2$ [Line 32].

Leaf agent $a_4$ initializes its (reduced) context $X^{a_4}$ to $\{(a_2, 0, 0)\}$ [Line 1]. It updates the lower bounds of nodes $i$ and $j$ to their delta costs 3 and 8, respectively [Line 19]. It updates the lower bound of node $E$ to 3 [Line 21]. It updates the upper bounds of nodes $i$ and $j$ to their delta costs 3 and 8, respectively [Line 20]. It updates the upper bound of node $E$ to 3 [Line 22]. The bounds of node $E$ that leaf agent $a_4$ maintains are not shown in the figure. It takes on its best value 0 [Line 15], initializes its ID to 1 [Lines 2 and 16], initializes its threshold to infinity [Line 17] and sends a COST message to its parent agent $a_2$ [Line 32].

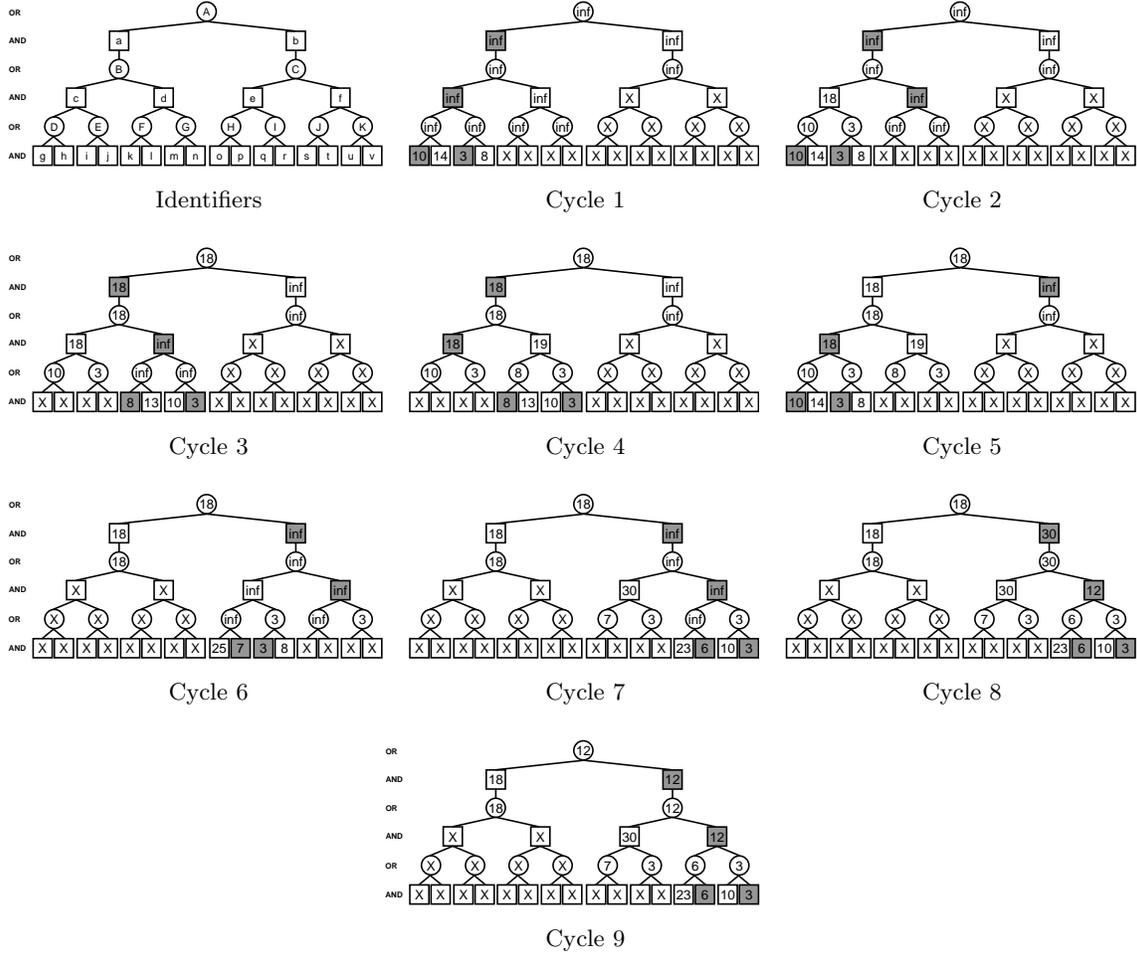

Figure 11: Trace of the Update of the Upper Bounds of BnB-ADOPT

In summary, the following messages are sent during Cycle 1:

- message (VALUE, $a_1$, 0, 1, $\infty$) from agent $a_1$ to agent $a_2$;

- message (VALUE, $a_1$, 0, 1, $\infty$) from agent $a_1$ to agent $a_3$;

- message (VALUE, $a_2$, 0, 1, $\infty$) from agent $a_2$ to agent $a_3$;

- message (VALUE, $a_2$, 0, 1, $\infty$) from agent $a_2$ to agent $a_4$;

- message (COST, $a_2$, $\{(a_1,0,0)\}$, 9, $\infty$) from agent $a_2$ to agent $a_1$;

- message (COST, $a_3$, $\{(a_1,0,0),(a_2,0,0)\}$, 10, 10) from agent $a_3$ to agent $a_2$; and

- message (COST, $a_4$, $\{(a_2,0,0)\}$, 3, 3) from agent $a_4$ to agent $a_2$.

- **Cycle 2:** Root agent $a_1$ receives the COST message sent by its child agent $a_2$ in Cycle 1. Since the context of agent $a_1$ ($= \{\}$) is compatible with the context in the message ($= \{(a_1,0,0)\}$), it improves its bounds. It updates the bounds of node $B$ to the bounds in the message ($= 9$ and infinity, respectively) [Lines 48-50]. It updates the bounds of nodes $a$, $b$ and $A$ [Lines 18-22].

It does not change its value since the lower bound of node $a$ ($= LB_{X'^{a_1}}^{a_1}(d^{a_1}) = 9$ for its value $d^{a_1} = 0$) is still smaller than its pruning quantity ($= \min\{TH_{X'^{a_1}}^{a_1}, UB_{X'^{a_1}}^{a_1}\} = \min(\infty, \infty) = \infty$). It sends VALUE messages to its child agent $a_2$ and pseudo-child agent $a_3$ [Lines 30-31].

Agent $a_2$ receives the VALUE message sent by its parent agent $a_1$ in Cycle 1. It updates its context from $\{(a_1, 0, 0)\}$ to $\{(a_1, 0, 1)\}$ since the ID of agent $a_1$ in its context ($= 0$) is smaller than the ID in the message ($= 1$) [Line 34]. Its threshold ($= \infty$) remains unchanged since it is the same as the desired threshold ($= \infty$) in the message. Agent $a_2$ also receives the COST messages sent by its child agents $a_3$ and $a_4$ in Cycle 1. Since its context ($= \{(a_1, 0, 1)\}$) is compatible with the contexts in the messages ($= \{(a_1, 0, 0), (a_2, 0, 0)\}$ and $\{(a_2, 0, 0)\}$, respectively), it improves its bounds. It updates the bounds of node $D$ to the bounds in the first message ($= 10$ and 10, respectively) and the bounds of node $E$ to the bounds in the second message ($= 3$ and 3, respectively) [Lines 48-50]. It updates the bounds of nodes $c$, $d$ and $B$ [Lines 18-22]. It changes its value since the lower bound of node $c$ ($= LB_{X'^{a_2}}^{a_2}(d^{a_2}) = 18$ for its value $d^{a_2} = 0$) is no longer smaller than its pruning quantity ($= \min\{TH_{X'^{a_2}}^{a_2}, UB_{X'^{a_2}}^{a_2}\} = \min(\infty, 18) = 18$). It takes on its best value 1 [Line 24], increments its ID to 2 [Lines 25-26], sends VALUE messages to its child agents $a_3$ and $a_4$ [Lines 30-31] and sends a COST message to its parent agent $a_1$ [Line 32].

Leaf agent $a_3$ receives the VALUE messages sent by its parent agent $a_2$ and pseudo-parent agent $a_1$ in Cycle 1. It updates its context from $\{(a_1, 0, 0), (a_2, 0, 0)\}$ to $\{(a_1, 0, 1), (a_2, 0, 1)\}$ since the IDs of agents $a_1$ and $a_2$ in its context ($= 0$ and 0, respectively) are smaller than the IDs in the messages ($= 1$ and 1, respectively) [Line 34]. Its threshold ($= \infty$) remains unchanged since it is the same as the desired threshold ($= \infty$) in the message. Its bounds are not reinitialized since its context is compatible with its previous context [Line 35]. It sends the same COST message as before to its parent agent $a_2$ [Line 32].

Leaf agent $a_4$ receives the VALUE message sent by its parent agent $a_2$ in Cycle 1. It updates its contexts from $\{(a_2, 0, 0)\}$ to $\{(a_2, 0, 1)\}$ since the ID of agent $a_2$ in its context ($= 0$) is smaller than the ID in the message ($= 1$) [Line 34]. Its threshold ($= \infty$) remains unchanged since it is the same as the desired threshold ($= \infty$) in the message. Its bounds are not reinitialized since its context is compatible with its previous context [Line 35]. It sends the same COST message as before to its parent agent $a_2$ [Line 32].

In summary, the messages sent during Cycle 2 are identical to the ones sent during Cycle 1, except for the messages sent by agents $a_2$, $a_3$ and $a_4$, which are as follows:

- message (VALUE, $a_2$, 1, 2, 8) from agent $a_2$ to agent $a_3$;
- message (VALUE, $a_2$, 1, 2, 8) from agent $a_2$ to agent $a_4$; and
- message (COST, $a_2$, $\{(a_1, 0, 1)\}$, 12, 18) from agent $a_2$ to agent $a_1$.
- message (COST, $a_3$, $\{(a_1, 0, 1), (a_2, 0, 1)\}$, 10, 10) from agent $a_3$ to agent $a_2$; and
- message (COST, $a_4$, $\{(a_2, 0, 1)\}$, 3, 3) from agent $a_4$ to agent $a_2$.

The VALUE messages are different because agent $a_2$ changed its value from 0 to 1. The COST messages are different because agent $a_2$ changed its bounds and its context and agents $a_3$ and $a_4$ changed their contexts.

- **Cycles 3-9:** The messages sent during Cycle 3 are identical to the ones sent during Cycle 2, except for the messages sent by agents $a_3$ and $a_4$, which are as follows:

  - message (COST, $a_3$, $\{(a_1, 0, 1), (a_2, 1, 2)\}$, 8, 8) from agent $a_3$ to agent $a_2$; and
  - message (COST, $a_4$, $\{(a_2, 1, 2)\}$, 3, 3) from agent $a_4$ to agent $a_2$.

The COST messages are different because agents $a_3$ and $a_4$ changed their contexts. The termination conditions holds after a finite amount of time when the upper bound of node $A$ ($= UB_{X^{a_1}}^{a_1} = 12$) is no larger than the lower bound of node $A$ ($= LB_{X^{a_1}}^{a_1} = 12$) [Line 27]. Root agent $a_1$ sends TERMINATE messages to all child agents [Line 28], and the TERMINATE messages propagate down the pseudo-tree [Line 28] until all agents terminate. BnB-ADOPT terminates after nine cycles with minimal solution cost 12.

# 4. Bounded-Error Approximations

In this section, we present three approximation mechanisms that allow BnB-ADOPT to trade off solution cost for a smaller runtime. They bound the error on the solution cost by a user-defined error bound. First, we modify the Absolute Error Mechanism of ADOPT (Modi et al., 2005) to work with BnB-ADOPT. This approximation mechanism allows users to specify an absolute error bound on the solution cost (for example, that the solution cost should be at most 10 larger than the minimal solution cost). However, it is often much more desirable to specify a relative error bound on the solution cost (for example, that the solution cost should be at most 10 percent larger than the minimal solution cost or, equivalently, 1.1 times larger than the minimal solution cost). This cannot be done with the Absolute Error Mechanism without knowing the minimal solution cost a priori. Thus, we introduce two approximation mechanisms that allow users to specify a relative error bound on the solution cost, namely the Relative Error Mechanism and the Weighted Heuristics Mechanism.

All approximation mechanisms let the root agent $r$ (and only the root agent) maintain the limit $lim^r$. The root agent uses this limit in the same way in the termination condition for all approximation mechanisms but updates it differently. The termination condition $UB_{X^r}^r \leq LB_{X^r}^r$ on Line 27 of the pseudocode of BnB-ADOPT is replaced with $UB_{X^r}^r \leq lim^r$. The root agent updates the limit between Lines 26 and 27 in the pseudocode, outside of the preceding if statement.

## 4.1 Absolute Error Mechanism

The Absolute Error Mechanism of ADOPT requires a user-defined absolute error bound $0 \leq b < \infty$ that specifies that the solution cost should be at most $b$ larger than the minimal solution cost. This approximation mechanism can easily be modified for BnB-ADOPT by setting the limit as follows:

$$lim^r \quad := \quad b + LB_{X^r}^r \tag{20}$$

BnB-ADOPT$_{AEM}$ is the resulting variant of BnB-ADOPT with the Absolute Error Mechanism. BnB-ADOPT$_{AEM}$ terminates once the upper bound of the root node (which is equal to the solution cost of the solution with the smallest solution cost found so far) is no larger than the limit (which is equal to the absolute error bound $b$ plus the lower bound of the root node, which is a lower bound on the minimal solution cost). BnB-ADOPT$_{AEM}$ terminates with a solution cost that is equal to the upper bound of the root node although the minimal solution cost could be as small as the lower bound of the root node. It thus terminates with a solution cost that is at most $b$ larger than the minimal solution cost. Figure 12 shows a trace of BnB-ADOPT$_{AEM}$ with absolute error bound $b = 24$ for our example DCOP problem. BnB-ADOPT$_{AEM}$ terminates after three cycles with suboptimal solution cost 18, which is six cycles faster than BnB-ADOPT.

## 4.2 Relative Error Mechanism

It is often much more desirable to specify a relative error bound on the solution cost rather than an absolute error bound. Fortunately, the Absolute Error Mechanism of BnB-ADOPT can easily be changed to the Relative Error Mechanism by setting the limit as follows. The Relative Error

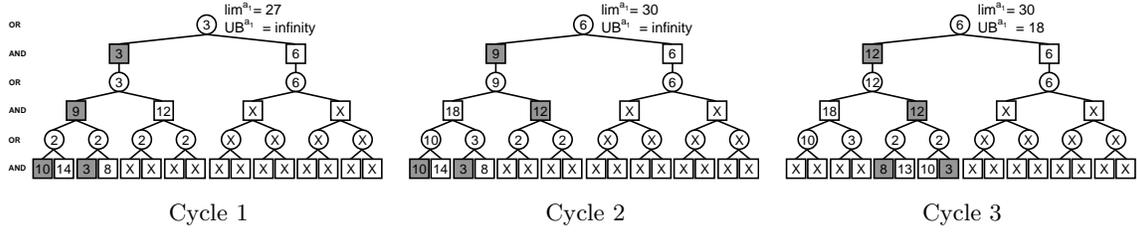

Figure 12: Trace of the Update of the Lower Bounds of BnB-ADOPT$_{AEM}$ with $b = 24$

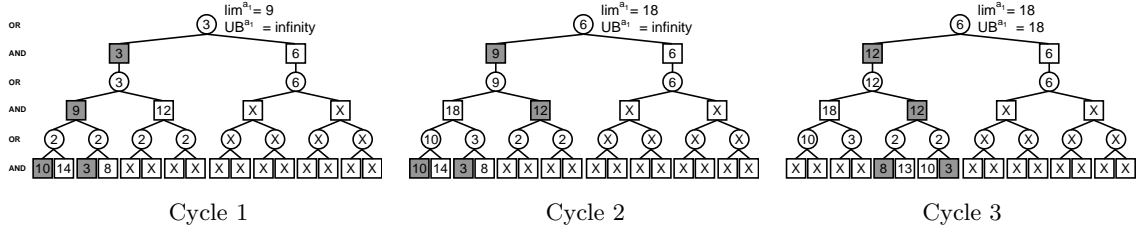

Figure 13: Trace of the Update of the Lower Bounds of BnB-ADOPT$_{REM}$ with $p = 3$

Mechanism requires a user-defined relative error bound $1 \leq p < \infty$ that specifies that the solution cost should be at most $p$ times larger than the minimal solution cost:

$$lim^r \quad := \quad p \cdot LB^r_{X^r} \tag{21}$$

BnB-ADOPT$_{REM}$ is the resulting variant of BnB-ADOPT with the Relative Error Mechanism. BnB-ADOPT$_{REM}$ terminates once the upper bound of the root node (which is equal to the solution cost of the solution with the smallest solution cost found so far) is no larger than the limit (which is equal to the relative error bound $p$ times the lower bound of the root node, which is a lower bound on the minimal solution cost). BnB-ADOPT$_{REM}$ terminates with a solution cost that is equal to the upper bound of the root node although the minimal solution cost could be as small as the lower bound of the root node. It thus terminates with a solution cost that is at most $p$ times larger than the minimal solution cost. Figure 13 shows a trace of BnB-ADOPT$_{REM}$ with relative error bound $p = 3$ for our example DCOP problem. BnB-ADOPT$_{REM}$ terminates after three cycles with suboptimal solution cost 18, which is six cycles faster than BnB-ADOPT.

### 4.3 Weighted Heuristics Mechanism

There is a second way of implementing a relative error bound for BnB-ADOPT since BnB-ADOPT uses admissible heuristic values. It is common practice in the context of A* to trade off solution cost for a smaller runtime by using weighted heuristic values (Pohl, 1973), which are derived from admissible heuristic values by multiplying them with a user-defined weight $1 \leq w < \infty$. The resulting heuristic values can be inadmissible. A* is then no longer guaranteed to find cost-minimal solutions but it is guaranteed to terminate with a solution cost that is at most $w$ times larger than the minimal solution cost (Pohl, 1970). This approximation mechanism can easily be modified for BnB-ADOPT by setting the limit as follows:

$$lim^r \quad := \quad LB^r_{X^r} \tag{22}$$

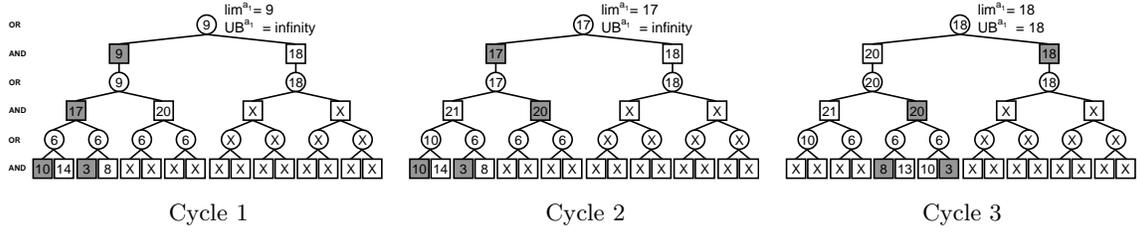

Figure 14: Trace of the Update of the Lower Bounds of BnB-ADOPT$_{WHM}$ with $w = 3$

and by initializing the lower bounds $lb_{X^a}^{a,c}(d)$ as follows:

$$lb_{X^a}^{a,c}(d) \quad := w \cdot h_{X^a}^{a,c}(d) \tag{23}$$

for all agents $a$, all values $d$, all child agents $c$ and all contexts $X^a$. BnB-ADOPT$_{WHM}$ is the resulting variant of BnB-ADOPT with the Weighted Heuristics Mechanism. BnB-ADOPT$_{WHM}$ terminates once the upper bound of the root node (which is equal to the solution cost of the solution with the smallest solution cost found so far) is no larger than the limit (which is equal to the lower bound of the root node, which is a lower bound on $w$ times the minimal solution cost). BnB-ADOPT$_{WHM}$ terminates with a solution cost that is equal to the upper bound of the root node although the minimal solution cost could be as small as the lower bound of the root node divided by $w$. It thus terminates with a solution cost that is at most $w$ times larger than the minimal solution cost. Figure 14 shows a trace of BnB-ADOPT$_{WHM}$ with $w = 3$ for our example DCOP problem. BnB-ADOPT$_{WHM}$ terminates after three cycles with suboptimal solution cost 18, which is six cycles faster than BnB-ADOPT.

## 5. Correctness and Completeness

In this section, we prove the correctness and completeness of BnB-ADOPT and its suboptimal variants. All definitions, lemmata, theorems and corollaries hold for BnB-ADOPT and its suboptimal variants except when mentioned otherwise. Therefore, each agent $a$ uses the following update equation for all values $d$, all child agents $c$ and all contexts $X^a$ to initialize its bounds $lb_{X^a}^{a,c}(d)$:

$$lb_{X^a}^{a,c}(d) := w \cdot h_{X^a}^{a,c}(d) \tag{24}$$

where the weight $w$ is a floating point number that satisfies $1 \leq w < \infty$ and the heuristic values $h_{X^a}^{a,c}(d)$ are floating point numbers that satisfy

$$0 \leq h_{X^a}^{a,c}(d) \leq \gamma_{X^a \cup (a,d)}^c \tag{25}$$

Messages are sent at the end of a cycle and received in the beginning of a cycle. $\Delta$ is the largest duration between the time a message is sent and the time it is processed, and $\epsilon$ is the largest duration of a cycle.

**Lemma 1.** *If two contexts $X$ and $X'$ of an arbitrary agent $a \in A$ agree on the values of all ancestor agents $p \in SCP(a)$ of agent $a$, then $\gamma_X^a = \gamma_{X'}^a$.*

*Proof.* By definition, $X^a \subseteq X$ is the (reduced) context that contains the values of all ancestor agents $p \in SCP(a)$ of agent $a$. The gamma cost $\gamma_X^a$ is the sum of the constraint costs of all constraints that involve agent $a$ or one of its descendant agents minimized over all possible values of agent $a$ and its descendant agents, under the assumption that the ancestor agents of agent $a$ take on the values in context $X$. The gamma cost $\gamma_X^a$ thus depends only on the values of the ancestor agents (including the parent agent) of agent $a$ that are the parent or pseudo-parent agents of agent $a$ or one (or more) of its descendant agents, that is, the values of all ancestor agents $p \in SCP(a)$ of agent $a$. Therefore, $\gamma_X^a = \gamma_{X^a}^a$. Similarly, $\gamma_{X'}^a = \gamma_{X^a}^a$. $\qquad\qquad\blacksquare$

**Definition 1.** *Contexts are correct iff the IDs of the values of all agents in the contexts are equal to the IDs of the agents, which implies that the values of all agents in the contexts are equal to the values of the agents.*

**Lemma 2.** *If the context $X^a$ of an arbitrary agent $a \in A$ does not change for a period of time, then the lower bounds $lb_{X^a}^{a,c}(d)$, $LB_{X^a}^a(d)$ and $LB_{X^a}^a$ are monotonically non-decreasing and the upper bounds $ub_{X^a}^{a,c}(d)$, $UB_{X^a}^a(d)$ and $UB_{X^a}^a$ are monotonically non-increasing during that period of time for all values $d \in Dom(a)$ and all child agents $c \in C(a)$.*

*Proof.* Since the context $X^a$ does not change, the delta values $\delta_{X^a}^a(d)$ are constant and the bounds (once initialized) are updated according to update equations 8 to 13. Thus, the lower bounds are monotonically non-decreasing and the upper bounds are monotonically non-increasing. $\qquad\qquad\blacksquare$

**Lemma 3.** *If the value of an arbitrary ancestor agent $p \in SCP(a)$ of an arbitrary agent $a \in A$ does not change between the current time $T$ and a future time $t$ with $t \geq T + |A| \cdot (\Delta + \epsilon) + \epsilon$, then the value of agent $p$ and its ID in the context of agent $a$ are equal to the value of agent $p$ and its ID, respectively, between some time $t'$ and time $t$ with $t' \leq t$.*

*Proof.* Assume that the value of an arbitrary ancestor agent $p \in SCP(a)$ of an arbitrary agent $a \in A$ does not change between the current time $T$ and a future time $t$ with $t \geq T + |A| \cdot (\Delta + \epsilon) + \epsilon$. There are the following two cases.

- Case 1: If agent $p$ is a parent or pseudo-parent agent of agent $a$, then it sent a VALUE message to agent $a$ with its value and ID at time $t'' \leq T + \epsilon$, that is, in the same cycle in which it took on the value that it has at time $T$ since the duration of that cycle is no larger than $\epsilon$. (The agents send VALUE messages at the end of every cycle.) Agent $a$ receives the VALUE message by time $t'' + \Delta$ since messages are delivered with finite delay $\Delta$. It then updates the value of agent $p$ and its ID in its context by time $t'' + \Delta$ since the update is done in the same cycle and the duration of that cycle is no larger than $\epsilon$. Thus, the value of agent $p$ and its ID in the context of agent $a$ are equal to the value of agent $p$ and its ID, respectively, between some time $t'$ and time $t$ with $t'' \leq t' \leq t'' + \Delta + \epsilon \leq T + \Delta + 2 \cdot \epsilon \leq t$ since agent $p$ does not change its value between time $t''$ and time $t$.

- Case 2: If agent $p$ is not a parent or pseudo-parent agent of agent $a$, then one of its pseudo-child agents $c$ is a descendant agent of agent $a$. Agent $p$ sent a VALUE message to agent $c$ with its value and ID at time $t'' \leq T + \epsilon$. Agent $c$ receives the VALUE message by time $t'' + \Delta$. It then updates the value of agent $p$ and its ID in its context and sends a COST message to its parent agent $pa(c)$ with its updated context by time $t'' + \Delta + \epsilon$. (The agents send COST messages at the end of every cycle.) Agent $pa(c)$ receives the COST message by time $t'' + 2 \cdot \Delta + \epsilon$. It then updates the value of agent $p$ and its ID in its context and sends a COST message to its parent agent $pa(pa(c))$ with its updated context by time $t'' + 2 \cdot (\Delta + \epsilon)$. This process continues until agent $a$ updates the value of agent $p$ and its ID in its context by time $t'' + n \cdot (\Delta + \epsilon)$, where $n \leq |A|$ is the number of messages in the chain of messages. Thus, the value of agent $p$ and its ID in the context of agent $a$ are equal to the value of agent $p$ and its ID, respectively, between some time $t'$ and time $t$ with $t'' \leq t' \leq t'' + n \cdot (\Delta + \epsilon) \leq T + |A| \cdot (\Delta + \epsilon) + \epsilon \leq t$ since agent $p$ does not change its value between time $t''$ and time $t$.

□

**Corollary 1.** *If the values of all ancestor agents $p \in SCP(a)$ of an arbitrary agent $a \in A$ do not change between the current time $T$ and a future time $t$ with $t \geq T + |A| \cdot (\Delta + \epsilon) + \epsilon$, then the context of agent $a$ is correct between some time $t'$ and time $t$ with $t' \leq t$.*

**Lemma 4.** *If $LB^c_{X^c} \leq w \cdot \gamma^c_{X^c} \leq w \cdot UB^c_{X^c}$ at all times for all child agents $c \in C(a)$ of an arbitrary agent $a$ and their contexts $X^c$, then $lb^{a,c}_{X^a}(d) \leq w \cdot \gamma^c_{X^a \cup (a,d)} \leq w \cdot ub^{a,c}_{X^a}(d)$ at all times for the context $X^a$ of agent $a$, all values $d \in Dom(a)$ and all child agents $c \in C(a)$.*

*Proof.* We prove the lemma by induction on the number of times that agent $a$ changes its context or updates its bounds $lb^{a,c}_{X^a}(d)$ and $ub^{a,c}_{X^a}(d)$ for an arbitrary value $d$ and an arbitrary child agent $c$ after agent $a$ initializes its bounds. The conclusion of the lemma holds after agent $a$ with context $X^a$ initializes its bounds since

$$lb^{a,c}_{X^a}(d) = w \cdot h^{a,c}_{X^a}(d) \qquad \text{(Eq. 24)}$$
$$\leq w \cdot \gamma^c_{X^a \cup (a,d)} \qquad \text{(Eq. 25)}$$
$$\leq \infty$$
$$= w \cdot ub^{a,c}_{X^a}(d) \qquad \text{(Eq. 7)}$$

for the (unchanged or new) context $X^a$ of agent $a$ (induction basis). Now assume that the lemma holds after agent $a$ changed its context or updated its bounds a number of times (induction assumption). We show that it then also holds after agent $a$ changes its context or updates its bounds one more time (induction step). There are the following two cases (where we split the operations after receiving a COST message into two parts).

- Case 1: The conclusion of the lemma holds when agent $a$ changes its context from $X^a$ to $\hat{X}^a$ after receiving a VALUE or COST message and the two contexts agree on the values of all ancestor agents $p \in SCP(c)$ since agent $a$ then does not change its bounds and thus

$$lb^{a,c}_{\hat{X}^a}(d) = lb^{a,c}_{X^a}(d) \qquad \text{(premise of case)}$$
$$\leq w \cdot \gamma^c_{X^a \cup (a,d)} \qquad \text{(induction assumption)}$$
$$= w \cdot \gamma^c_{\hat{X}^a \cup (a,d)} \qquad \text{(Lemma 1)}$$
$$ub^{a,c}_{\hat{X}^a}(d) = ub^{a,c}_{X^a}(d) \qquad \text{(premise of case)}$$
$$\geq \gamma^c_{X^a \cup (a,d)} \qquad \text{(induction assumption)}$$
$$= \gamma^c_{\hat{X}^a \cup (a,d)} \qquad \text{(Lemma 1)}$$

after receiving the VALUE or COST message (since contexts $X^a$ and $\hat{X}^a$ agree on the values of all ancestor agents $p \in SCP(c)$).

- Case 2: The conclusion of the lemma holds when agent $a$ updates its bounds from $lb^{a,c}_{X^a}(d)$ and $ub^{a,c}_{X^a}(d)$ to $\hat{lb}^{a,c}_{X^a}(d)$ and $\hat{ub}^{a,c}_{X^a}(d)$, respectively, after receiving a COST message from some child agent $c$ with bounds $LB^c_{X^c}$ and $UB^c_{X^c}$ and context $X^c$ that is compatible with its context $X^a$ and in which agent $a$ has value $d$ since

$$\hat{lb}^{a,c}_{X^a}(d) = \max\{lb^{a,c}_{X^a}(d), LB^c_{X^c}\} \qquad \text{(Eq. 8)}$$

$$\leq \max\{w \cdot \gamma^c_{X^a \cup (a,d)}, w \cdot \gamma^c_{X^c}\} \qquad \text{(induction assumption and premise of lemma)}$$

$$= \max\{w \cdot \gamma^c_{X^a \cup (a,d)}, w \cdot \gamma^c_{X^a \cup (a,d)}\} \qquad \text{(Lemma 1)}$$

$$= w \cdot \gamma^c_{X^a \cup (a,d)}$$

$$\hat{ub}^{a,c}_{X^a}(d) = \min\{ub^{a,c}_{X^a}(d), UB^c_{X^c}\} \qquad \text{(Eq. 11)}$$

$$\geq \min\{\gamma^c_{X^a \cup (a,d)}, \gamma^c_{X^c}\} \qquad \text{(induction assumption and premise of lemma)}$$

$$= \min\{\gamma^c_{X^a \cup (a,d)}, \gamma^c_{X^a \cup (a,d)}\} \qquad \text{(Lemma 1)}$$

$$= \gamma^c_{X^a \cup (a,d)}$$

after receiving the COST message (since contexts $X^a \cup (a,d)$ and $X^c$ agree on the values of all ancestor agents $p \in SCP(c)$).

Thus, $lb^{a,c}_{X^a}(d) \leq w \cdot \gamma^c_{X^a \cup (a,d)} \leq w \cdot ub^{a,c}_{X^a}(d)$ at all times for all values $d \in Dom(a)$ and all child agents $c \in C(a)$. $\qquad \square$

**Lemma 5.** $LB^a_{X^a}(d) \leq w \cdot \gamma^a_{X^a}(d) \leq w \cdot UB^a_{X^a}(d)$ and $LB^a_{X^a} \leq w \cdot \gamma^a_{X^a} \leq w \cdot UB^a_{X^a}$ at all times for all values $d \in Dom(a)$ and the context $X^a$ of an arbitrary agent $a \in A$.

*Proof.* We prove the lemma by induction on the depth of an agent in the pseudo-tree. The lemma holds for a leaf agent $a$ in the pseudo-tree with context $X^a$ since

$$LB^a_{X^a}(d) = \delta^a_{X^a}(d) \qquad \text{(Eq. 9)}$$

$$= \gamma^a_{X^a}(d) \qquad \text{(Eq. 1)}$$

$$UB^a_{X^a}(d) = \delta^a_{X^a}(d) \qquad \text{(Eq. 12)}$$

$$= \gamma^a_{X^a}(d) \qquad \text{(Eq. 1)}$$

for all values $d$ at all times. Thus, $LB^a_{X^a}(d) = \gamma^a_{X^a}(d) \leq w \cdot \gamma^a_{X^a}(d) = w \cdot UB^a_{X^a}(d)$ for all values $d$ at all times. Furthermore,

$$LB^a_{X^a} = \min_{d \in Dom(a)}\{LB^a_{X^a}(d)\} \qquad \text{(Eq. 10)}$$

$$= \min_{d \in Dom(a)}\{\gamma^a_{X^a}(d)\} \qquad \text{(see above)}$$

$$= \gamma^a_{X^a} \qquad \text{(Eq. 2)}$$

$$UB^a_{X^a} = \min_{d \in Dom(a)}\{UB^a_{X^a}(d)\} \qquad \text{(Eq. 13)}$$

$$= \min_{d \in Dom(a)}\{\gamma^a_{X^a}(d)\} \qquad \text{(see above)}$$

$$= \gamma^a_{X^a} \qquad \text{(Eq. 2)}$$

at all times. Thus, $LB^a_{X^a} = \gamma^a_{X^a} \leq w \cdot \gamma^a_{X^a} = w \cdot UB^a_{X^a}$ at all times (induction basis). Now assume that the lemma holds for all agents of depth $d$ in the pseudo-tree (induction assumption). We show that it then also holds for all agents of depth $d-1$ in the pseudo-tree each time after they update their bounds (induction step). The lemma holds for agent $a$ with context $X^a$ since

$$LB^a_{X^a}(d) = \delta^a_{X^a}(d) + \sum_{c \in C(a)} lb^{a,c}_{X^a}(d) \qquad \text{(Eq. 9)}$$

$$\leq \delta^a_{X^a}(d) + \sum_{c \in C(a)} w \cdot \gamma^c_{X^a \cup (a,d)} \qquad \text{(induction assumption and Lemma 4)}$$

$$\leq w \cdot \gamma^a_{X^a}(d) \qquad \text{(Eq. 1)}$$

$$UB^a_{X^a}(d) = \delta^a_{X^a}(d) + \sum_{c \in C(a)} ub^{a,c}_{X^a}(d) \qquad \text{(Eq. 12)}$$

$$\geq \delta^a_{X^a}(d) + \sum_{c \in C(a)} \gamma^c_{X^a \cup (a,d)} \qquad \text{(induction assumption and Lemma 4)}$$

$$= \gamma^a_{X^a}(d) \qquad \text{(Eq. 1)}$$

Thus, $LB^a_{X^a}(d) \leq w \cdot \gamma^a_{X^a}(d) \leq w \cdot UB^a_{X^a}(d)$ at all times for all values $d \in Dom(a)$. Furthermore,

$$LB^a_{X^a} = \min_{d \in Dom(a)} \{LB^a_{X^a}(d)\} \qquad \text{(Eq. 10)}$$

$$\leq \min_{d \in Dom(a)} \{w \cdot \gamma^a_{X^a}(d)\} \qquad \text{(see above)}$$

$$= w \cdot \min_{d \in Dom(a)} \{\gamma^a_{X^a}(d)\}$$

$$= w \cdot \gamma^a_{X^a} \qquad \text{(Eq. 2)}$$

$$UB^a_{X^a} = \min_{d \in Dom(a)} \{UB^a_{X^a}(d)\} \qquad \text{(Eq. 13)}$$

$$\geq \min_{d \in Dom(a)} \{\gamma^a_{X^a}(d)\} \qquad \text{(see above)}$$

$$= \gamma^a_{X^a} \qquad \text{(Eq. 2)}$$

Thus, $LB^a_{X^a} \leq w \cdot \gamma^a_{X^a} \leq w \cdot UB^a_{X^a}$ at all times. $\qquad \blacksquare$

**Definition 2.** *The potential of an agent $a \in A$ with context $X^a$ is $\sum_{d \in Dom(a)} \{w \cdot UB^a_{X^a}(d) - LB^a_{X^a}(d)\}$.*

**Lemma 6.** *If the context $X^a$ of an arbitrary agent $a \in A$ no longer changes, then the potential of the agent is monotonically non-increasing and decreases by more than a positive constant every time the agent changes its value.*

*Proof.* The lower bounds $LB^a_{X^a}(d)$ are monotonically non-decreasing and the upper bounds $UB^a_{X^a}(d)$ are monotonically non-increasing for all values $d$ according to Lemma 2 since the context $X^a$ of agent $a$ no longer changes. Therefore, the potential of agent $a$ is monotonically non-increasing. Furthermore, agent $a$ changes its value $d$ to a new value only if $\min_{d \in Dom(a)} \{LB^a_{X^a}(d)\} < LB^a_{X^a}(d)$ [Line 24]. Thus, the lower bound $LB^a_{X^a}(d)$ must have strictly increased between the time when the agent changed its value to $d$ and the time when it changes its value $d$ to a new value. Thus, its potential has decreased by more than a positive constant, namely the smallest possible increase of the lower bound $LB^a_{X^a}(d)$. Assume that all constraint costs, weights and heuristic values are integers. Then, the smallest possible increase is bounded from below by one because the only possible values of $LB^a_{X^a}(d)$ are combinations of all constraint costs and weighted heuristic values. A similar statement holds if all constraint costs, weights and heuristic values are floating point numbers since they can then all be transformed into integers by multiplying them with the same sufficiently large integer. $\qquad \blacksquare$

**Lemma 7.** *All agents change their values only a finite number of times.*

*Proof.* Assume that the lemma does not hold and choose an agent $a$ that changes its value an infinite number of times but whose ancestor agents $p \in SCP(a)$ change their values only a finite number of times. Then, there exists a time when the ancestor agents do not change their values any longer. There exists a (later) time when agent $a$ no longer changes its context $X^a$ according to Corollary 1. Every time agent $a$ changes its value afterwards, its potential decreases by more than a positive constant according to Lemma 6, towards minus infinity. However, its potential cannot become negative since $LB^a_{X^a}(d) \leq w \cdot UB^a_{X^a}(d)$ for all values $d$ according to Lemma 5, which is a contradiction. Thus, all agents change their values only a finite number of times. □

**Lemma 8.** *If BnB-ADOPT and its suboptimal variants do not terminate earlier, then* $UB^a_{X^a} \leq LB^a_{X^a}$ *after a finite amount of time for all agents* $a \in A$ *and their contexts* $X^a$.

*Proof.* We prove the lemma by induction on the depth of an agent in the pseudo-tree. There exists a time when no agent changes its value any longer according to Lemma 7. There exists a (later) time when the contexts of all agents are correct and no longer change according to Corollary 1. Let $X^a$ be the context of agent $a$ at this point in time for all agents $a$. There exists an (even later) time when the bounds $lb^{a,c}_{X^a}(d)$, $LB^a_{X^a}(d)$, $LB^a_{X^a}$, $ub^{a,c}_{X^a}(d)$, $UB^a_{X^a}(d)$ and $UB^a_{X^a}$ no longer change for all agents $a$, all values $d$ and all child agents $c$ since (1) the lower bounds $lb^{a,c}_{X^a}(d)$, $LB^a_{X^a}(d)$ and $LB^a_{X^a}$ are monotonically non-decreasing and the upper bounds $lb^{a,c}_{X^a}(d)$, $UB^a_{X^a}(d)$ and $UB^a_{X^a}$ are monotonically non-increasing for all agents $a$, all values $d$ and all child agents $c$ according to Lemma 2, (2) $LB^a_{X^a}(d) \leq w \cdot \gamma^a_{X^a}(d) \leq w \cdot UB^a_{X^a}(d)$ and $LB^a_{X^a} \leq w \cdot \gamma^a_{X^a} \leq w \cdot UB^a_{X^a}$ for all agents $a$ and all values $d$ according to Lemma 5, (3) $lb^{a,c}_{X^a}(d) \leq w \cdot ub^{a,c}_{X^a}(d)$ for all agents $a$, all values $d$ and all child agents $c$ according to Lemma 4 and (4) the smallest possible increases of the lower bounds and the smallest possible decreases of the upper bounds are larger than a positive constant since the only possible values of the bounds are combinations of all constraint costs and heuristic values, as explained in more detail in the proof of Lemma 6. Consider the first COST message that each agent sends after this time and the earliest time when all of these COST messages have been processed by their receiving agents. The lemma holds for a leaf agent $a$ in the pseudo-tree with context $X^a$ since

$$LB^a_{X^a}(d) = \delta^a_{X^a}(d) \qquad \text{(Eq. 9)}$$
$$= \gamma^a_{X^a}(d) \qquad \text{(Eq. 1)}$$
$$UB^a_{X^a}(d) = \delta^a_{X^a}(d) \qquad \text{(Eq. 12)}$$
$$= \gamma^a_{X^a}(d) \qquad \text{(Eq. 1)}$$

for all values $d$ after the considered time. Furthermore,

$$LB^a_{X^a} = \min_{d \in Dom(a)} \{LB^a_{X^a}(d)\} \qquad \text{(Eq. 10)}$$
$$= \min_{d \in Dom(a)} \{\gamma^a_{X^a}(d)\} \qquad \text{(see above)}$$
$$= \gamma^a_{X^a} \qquad \text{(Eq. 2)}$$
$$UB^a_{X^a} = \min_{d \in Dom(a)} \{UB^a_{X^a}(d)\} \qquad \text{(Eq. 13)}$$
$$= \min_{d \in Dom(a)} \{\gamma^a_{X^a}(d)\} \qquad \text{(see above)}$$
$$= \gamma^a_{X^a} \qquad \text{(Eq. 2)}$$

after the considered time. Thus, $UB^a_{X^a} = LB^a_{X^a}$ after the considered time (induction basis). Now assume that the lemma holds for all agents of depth $d$ in the pseudo-tree after the considered time (induction assumption). We show that it then also holds for all agents of depth $d - 1$ in the pseudo-tree after the considered time (induction step). For agent $a$ with context $X^a$

$$LB_{X^a}^a(d) = \delta_{X^a}^a(d) + \sum_{c \in C(a)} lb_{X^a}^{a,c}(d) \qquad \text{(Eq. 9)}$$

$$= \delta_{X^a}^a(d) + \sum_{c \in C(a)} \max\{lb_{X^a}^{a,c}(d), LB_{X^c}^c\} \qquad \text{(Eq. 8)}$$

$$\geq \delta_{X^a}^a(d) + \sum_{c \in C(a)} LB_{X^c}^c$$

$$\geq \delta_{X^a}^a(d) + \sum_{c \in C(a)} UB_{X^c}^c \qquad \text{(induction assumption)}$$

$$\geq \delta_{X^a}^a(d) + \sum_{c \in C(a)} \min\{ub_{X^a}^{a,c}(d), UB_{X^c}^c\}$$

$$= \delta_{X^a}^a(d) + \sum_{c \in C(a)} ub_{X^a}^{a,c}(d) \qquad \text{(Eq. 11)}$$

$$= UB_{X^a}^a(d) \qquad \text{(Eq. 12)}$$

for its value $d$ after the considered time since all bounds no longer change. Thus, $UB_{X^a}^a(d) \leq LB_{X^a}^a(d)$ for its value $d$ after the considered time. Since agent $a$ does not change its value $d$ after the considered time, it must hold that $LB_{X^a}^a(d) < \min\{TH_{X^a}^a, UB_{X^a}^a\}$ [Line 23] or $LB_{X^a}^a(d) = \min_{d \in Dom(a)}\{LB_{X^a}^a(d)\}$ [Line 24]. The first disjunct implies that

$$\min\{TH_{X^a}^a, UB_{X^a}^a\} \leq UB_{X^a}^a$$

$$\leq UB_{X^a}^a(d) \qquad \text{(Eq. 13)}$$

$$\leq LB_{X^a}^a(d) \qquad \text{(see above)}$$

$$< \min\{TH_{X^a}^a, UB_{X^a}^a\} \qquad \text{(first disjunct)}$$

for its value $d$, which is a contradiction. The second disjunct implies that

$$UB_{X^a}^a \leq UB_{X^a}^a(d) \qquad \text{(Eq. 13)}$$

$$\leq LB_{X^a}^a(d) \qquad \text{(see above)}$$

$$= \min_{d \in Dom(a)}\{LB_{X^a}^a(d)\} \qquad \text{(second disjunct)}$$

$$= LB_{X^a}^a \qquad \text{(Eq. 10)}$$

for its value $d$ and thus that $UB_{X^a}^a \leq LB_{X^a}^a$. □

**Theorem 1.** *BnB-ADOPT and its suboptimal variants terminate after a finite amount of time.*

*Proof.* If BnB-ADOPT and its suboptimal variants do not terminate earlier, then $UB_{X^a}^a \leq LB_{X^a}^a$ after a finite amount of time for all agents $a \in A$ and their contexts $X^a$ according to Lemma 8. In particular, $UB_{X^r}^r \leq LB_{X^r}^r \leq lim^r$ for the root agent $r$, where $lim^r = LB_{X^r}^r$ for BnB-ADOPT and BnB-ADOPT$_{WHM}$, $lim^r = b + LB_{X^r}^r$ with $b \geq 0$ for BnB-ADOPT$_{AEM}$ and $lim^r = p \cdot LB_{X^r}^r$ with $p \geq 1$ for BnB-ADOPT$_{REM}$ according to Section 4. Thus, both the termination condition $UB_{X^r}^r \leq LB_{X^r}^r$ of BnB-ADOPT and the termination condition $UB_{X^r}^r \leq lim^r$ of its suboptimal variants are satisfied. □

**Theorem 2.** *BnB-ADOPT terminates with the minimal solution cost $\gamma_{X^r}^r$.*

*Proof.* BnB-ADOPT terminates after a finite amount of time according to Theorem 1. The solution cost of BnB-ADOPT is the upper bound $UB_{X^r}^r$ of the root agent $r$. $UB_{X^r}^r \leq LB_{X^r}^r$ upon termination according to its termination condition. $w \cdot UB_{X^r}^r \geq w \cdot \gamma_{X^r}^r \geq LB_{X^r}^r$ according to Lemma 5. Therefore, $UB_{X^r}^r = \gamma_{X^r}^r = LB_{X^r}^r$ since $w = 1$. $\qquad\blacksquare$

**Theorem 3.** *BnB-ADOPT$_{AEM}$ terminates with a solution cost that is bounded from above by the user-defined absolute error bound $b$ plus the minimal solution cost $\gamma_{X^r}^r$.*

*Proof.* BnB-ADOPT$_{AEM}$ terminates after a finite amount of time according to Theorem 1. The solution cost of BnB-ADOPT$_{AEM}$ is the upper bound $UB_{X^r}^r$ of the root agent $r$. $UB_{X^r}^r \leq lim^r = b + LB_{X^r}^r$ upon termination according to its termination condition. $LB_{X^r}^r \leq w \cdot \gamma_{X^r}^r$ according to Lemma 5. Therefore, $UB_{X^r}^r \leq b + \gamma_{X^r}^r$ since $w = 1$. $\qquad\blacksquare$

**Theorem 4.** *BnB-ADOPT$_{REM}$ terminates with a solution cost that is bounded from above by the user-defined relative error bound $p$ times the minimal solution cost $\gamma_{X^r}^r$.*

*Proof.* BnB-ADOPT$_{REM}$ terminates after a finite amount of time according to Theorem 1. The solution cost of BnB-ADOPT$_{REM}$ is the upper bound $UB_{X^r}^r$ of the root agent $r$. $UB_{X^r}^r \leq lim^r = p \cdot LB_{X^r}^r$ upon termination according to its termination condition. $LB_{X^r}^r \leq w \cdot \gamma_{X^r}^r$ according to Lemma 5. Therefore, $UB_{X^r}^r \leq p \cdot \gamma_{X^r}^r$ since $w = 1$. $\qquad\blacksquare$

**Theorem 5.** *BnB-ADOPT$_{WHM}$ terminates with a solution cost that is bounded from above by the user-defined weight $w$ times the minimal solution cost $\gamma_{X^r}^r$.*

*Proof.* BnB-ADOPT$_{WHM}$ terminates after a finite amount of time according to Theorem 1. The solution cost of BnB-ADOPT$_{WHM}$ is the upper bound $UB_{X^r}^r$ of the root agent $r$. $UB_{X^r}^r \leq lim^r = LB_{X^r}^r$ upon termination according to its termination condition. $LB_{X^r}^r \leq w \cdot \gamma_{X^r}^r$ according to Lemma 5. Therefore, $UB_{X^r}^r \leq w \cdot \gamma_{X^r}^r$. $\qquad\blacksquare$

## 6. Experimental Evaluations

In this section, we compare BnB-ADOPT to two other memory-bounded DCOP search algorithms that also restrict communication to agents that share constraints, namely ADOPT and NCBB. We also compare the three suboptimal variants of BnB-ADOPT to each other. We use the distributed DFS algorithm with the max-degree heuristic (Hamadi, Bessière, & Quinqueton, 1998) that is used by ADOPT to construct the pseudo-trees. We use DP2 (Ali et al., 2005) that is used by ADOPT to pre-calculate the heuristic values for ADOPT and BnB-ADOPT. DP2 solves a relaxed version of the given DCOP problem (where backedges are ignored) with a dynamic programming based approach. NCBB calculates its own heuristic values during the search rather than in a pre-processing step.

### 6.1 Runtime Metrics

We use two common runtime metrics, namely non-concurrent constraint checks (Meisels, Kaplansky, Razgon, & Zivan, 2002) and cycles (Modi et al., 2005).

- **Non-concurrent constraint checks (NCCCs):** NCCCs are a weighted sum of processing and communication time. Every agent $a$ maintains a counter $NCCC^a$, which is initialized to 0. The agent assigns $NCCC^a := NCCC^a + 1$ every time it performs a constraint check to account for the time it takes to perform the constraint check. It assigns $NCCC^a := \max\{NCCC^a, NCCC^{a'} + t\}$ every time it receives a message from agent $a'$ to account for the time it takes to wait for agent $a'$ to send the message ($NCCC^{a'}$) and the transmission time of the message ($t$). We use $t = 0$ to simulate fast communication and $t = 1000$ to simulate slow communication. The number of NCCCs then is the largest counter value of any agent.

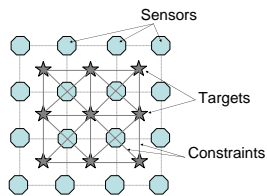

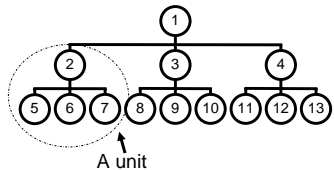

Figure 15: Example: Allocating Targets    Figure 16: Example: Scheduling Meetings

NCCCs are a good runtime metric if the ratio of processing and communication time can be estimated reliably.

- **Cycles:** Cycles are time slices. A cycle is the time required for an agent to process all incoming messages in its queue and send all outgoing messages, which are then processed by the receiving agents in the next cycle. Thus, the number of cycles indicates the length of the longest chain of messages between agents. Cycles are a good runtime metric if the communication time is much larger than the processing time. Cycles will become a better and better runtime metric in the future since the communication time is expected to remain relatively stable while the processing time is expected to decrease (Silaghi, Lass, Sultanik, Regli, Matsui, & Yokoo, 2008).

## 6.2 DCOP Problem Types

We use three DCOP problem types in our experiments, namely graph coloring problems, sensor network problems and meeting scheduling problems.

- **Graph coloring:** Graph coloring problems involve coloring the vertices of a graph, taking restrictions between the colors of adjacent vertices into account. The agents are the vertices, their domains are the colors, and the constraints are between adjacent vertices. We vary the number of vertices from 5 to 15, the constraint density (= the ratio between the number of constraints and the number of agents) from 2 (sparse graphs) to 3 (dense graphs) and the range of constraint costs from a range of 0 to 1 (small range) to a range of 0 to 10,000 (large range). Each agent always has three possible values. We average the experimental results over 50 DCOP problem instances with randomly generated constraints and randomly generated integer constraint costs.

- **Sensor network:** Sensor network problems involve assigning targets to sensors in a sensor network, taking restrictions in the availability of the sensors, restrictions in the number of sensors that need to track each target and the priorities of the targets into account. The agents are the targets, their domains are the time slots when they can be tracked, and the constraints are between adjacent targets (Maheswaran et al., 2004b). Figure 15 shows a sensor network where the targets are located on a grid and each target is surrounded by four sensors, all of which are needed to track the target. We vary the number of targets from 4 to 15. We always use 8 time slots. The cost of assigning a time slot to a target that is also assigned to an adjacent target is infinity (to be precise: 1,000,000) since the same sensor cannot track both targets during the same time slot. The cost of targets that are not tracked during any time slot is 100. All other costs are in the range of 0 to 100. We average the experimental results over 50 DCOP problem instances with randomly generated integer constraint costs.

- **Meeting scheduling:** Meeting scheduling problems involve scheduling meetings between the employees of a company, taking restrictions in their availability as well as their priorities into account. The agents are the meetings, their domains are the time slots when they can be held, and the constraints are between meetings that share participants (Maheswaran et al.,

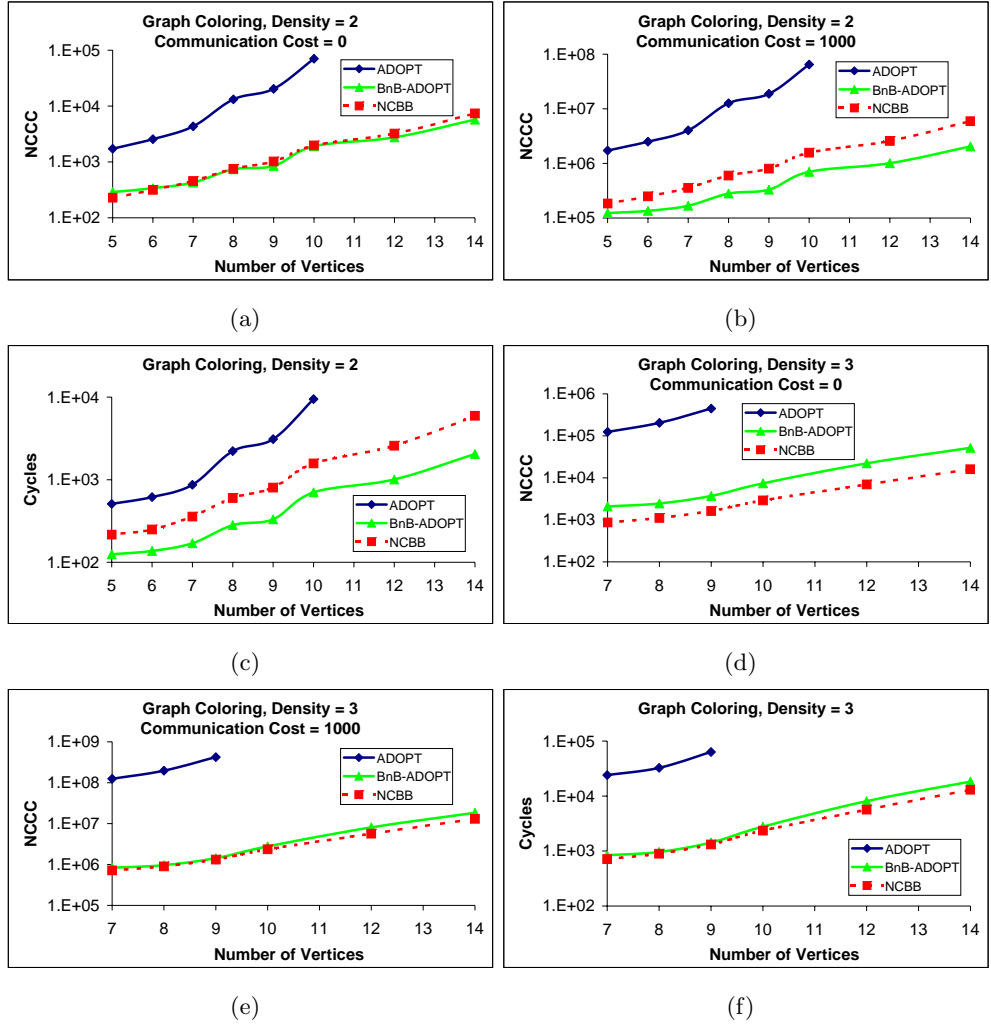

Figure 17: Experimental Results Comparing ADOPT, BnB-ADOPT and NCBB for Graph Coloring Problems with Constraint Costs Ranging from 0 to 10,000

2004b). Figure 16 shows a hierarchical organization with 4 units of a supervisor and its three subordinates. For example, supervisor 2 has three subordinates 5, 6 and 7. In each unit, we assume five possible meetings: one of the entire unit (e.g., 2, 5, 6, 7), two parent-child meetings (e.g., 2, 5 and 2, 7) and two sibling-sibling meetings (e.g., 5, 6 and 6, 7). We vary the number of meetings from 5 (1 unit) to 20 (4 units). We always use 8 time slots. The cost of assigning a time slot to a meeting that has at least one participant who has another meeting during the same time slot is infinity (to be precise: 1,000,000) since the same person cannot attend more than one meeting at a time. The cost of a non-scheduled meeting is 100. All other costs are in the range of 0 to 100. We average the experimental results over 50 DCOP problem instances with randomly generated integer constraint costs.

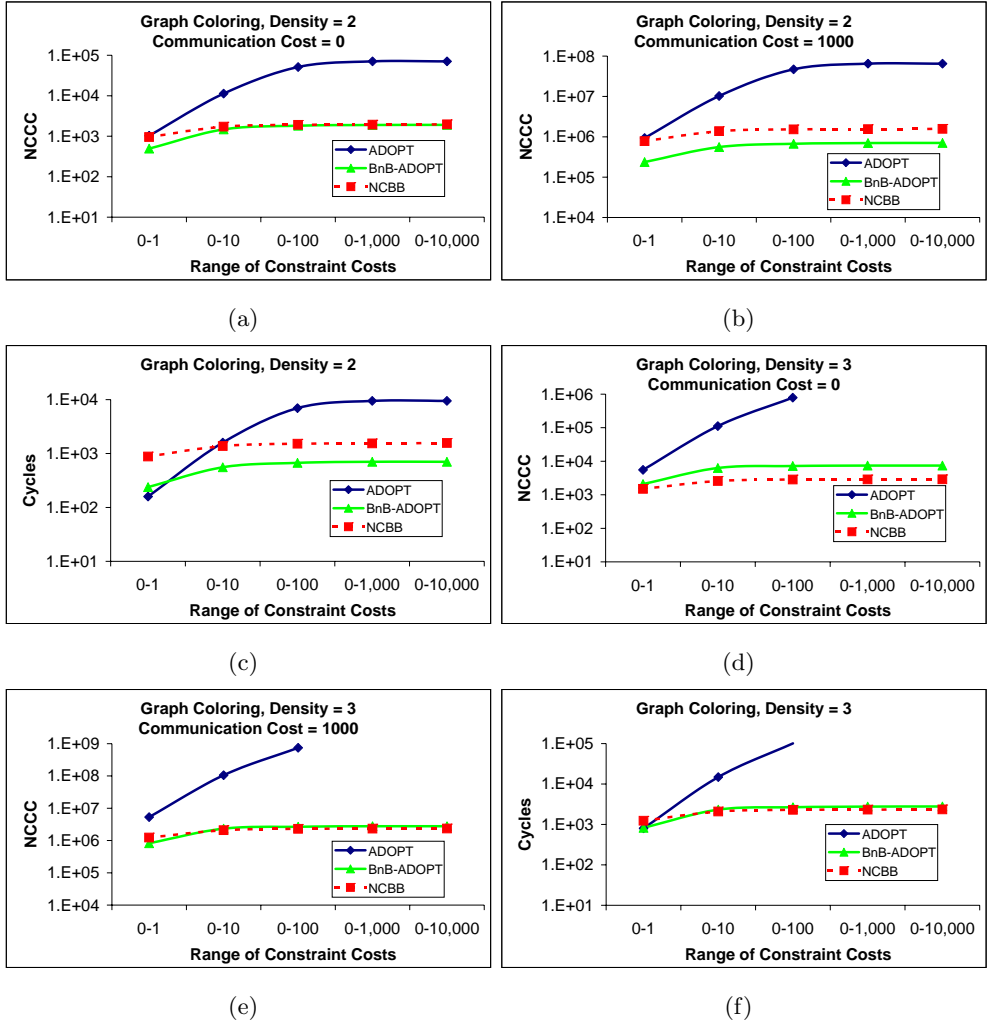

Figure 18: Experimental Results Comparing ADOPT, BnB-ADOPT and NCBB for Graph Coloring Problems with 10 Vertices

## 6.3 Experimental Results: Optimal DCOP Search Algorithms

We first compare BnB-ADOPT to ADOPT and NCBB. Figure 17 shows our experimental results for graph coloring problems with constraint costs ranging from 0 to 10,000, where we varied the number of vertices, while Figure 18 shows our experimental results for graph coloring problems with 10 vertices, where we varied the range of constraint costs. Figures 17(a-c) and 18(a-c) show the results for coloring sparse graphs, and Figures 17(d-f) and 18(d-f) show the results for coloring dense graphs. The y-axes are in log scale and show the runtimes in NCCCs or cycles. DCOP search algorithms on sparse graphs are faster than on dense graphs because, for example, there is a larger likelihood of independent DCOP subproblems in sparse graphs. BnB-ADOPT is generally faster than NCBB on sparse graphs but not on dense graphs because BnB-ADOPT allows agents to send messages only to their parent agents in the pseudo-tree (along edges of the pseudo-tree) but NCBB allows agents also to send messages to their pseudo-parent agents (along backedges of the pseudo-tree). Thus, agents of NCBB receive updates faster than agents of BnB-ADOPT. This effect is more prevalent in dense graphs since there are more backedges in dense graphs. However, the difference between BnB-ADOPT and NCBB becomes negligible when communication is slow.

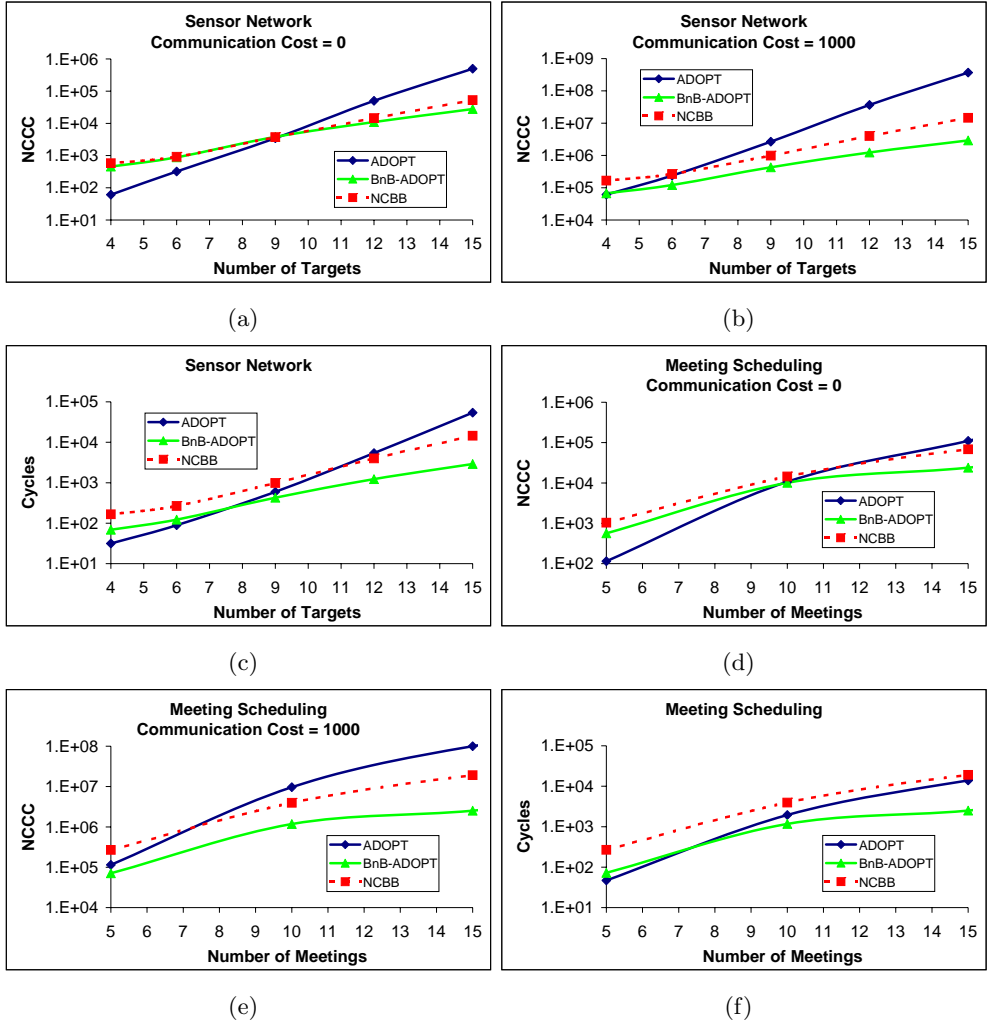

Figure 19: Experimental Results Comparing ADOPT, BnB-ADOPT and NCBB for Sensor Network and Meeting Scheduling Problems

Figure 17 shows that BnB-ADOPT is at least half an order of magnitude faster than ADOPT when the number of vertices is small. The speedup over ADOPT increases as the number of vertices gets larger and the DCOP problems thus become more complex. Similarly, Figure 18 shows that the speedup over ADOPT increases as the range of constant costs increases and the DCOP problems thus become more complex. However, ADOPT can be faster than BnB-ADOPT for simple DCOP problems. For example, ADOPT requires fewer cycles than BnB-ADOPT for DCOP problems with constraint costs ranging from 0 to 1. Figure 19 shows the same trend for sensor network and meeting scheduling problems. The reason for this behavior is as follows. ADOPT uses memory-bounded best-first search and thus exploits the heuristic values well but needs to repeatedly reconstruct partial solutions that it purged from memory, especially if the heuristic values are poorly informed. BnB-ADOPT uses depth-first branch-and-bound search and thus does not exploit the heuristic values quite as well but does not have to repeatedly reconstruct partial solutions. ADOPT can thus be faster than BnB-ADOPT for DCOP problems with well informed heuristic values, such as simple DCOP problems.

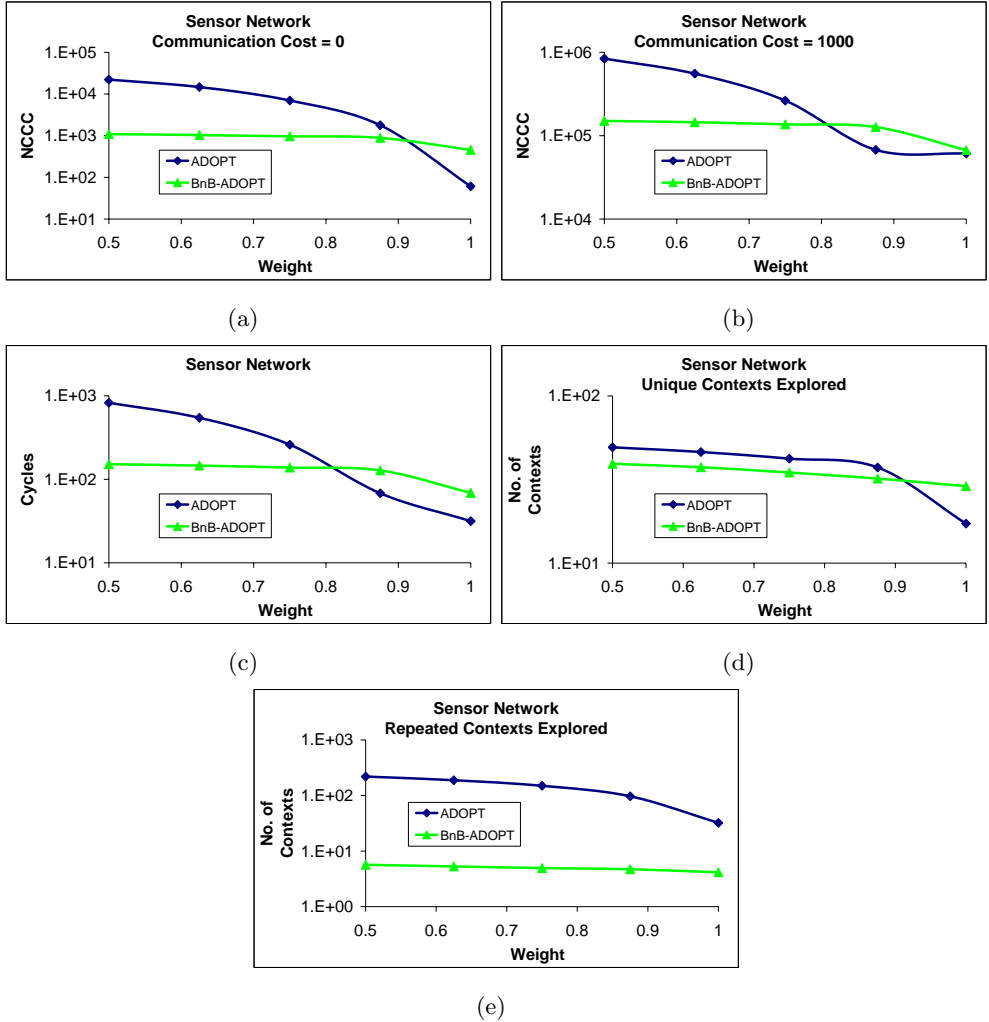

Figure 20: Experimental Results on the Cause of Speedup for ADOPT and BnB-ADOPT

We confirm this intuition with an additional experiment on sensor network problems with four targets and different informedness of heuristic values. We use the heuristic values $c \cdot h_{X^a}^{a,c}(d)$ for $0.5 \leq c \leq 1$, where $h_{X^a}^{a,c}(d)$ are the heuristic values calculated by DP2, as used until now. Figures 20(a-c) show the number of NCCCs for different weights $c$. When the heuristic values are well informed (large weights), ADOPT can indeed be faster than BnB-ADOPT. Since ADOPT relies on the heuristic values more than BnB-ADOPT, the speedup of ADOPT is much larger than that of BnB-ADOPT as the heuristic values get more informed. Figures 20(d) and 20(e) show the number of unique (= different) and repeated contexts per agent for different weights $c$. When the heuristic values are well informed (large weights), agents of ADOPT explore fewer unique contexts than agents of BnB-ADOPT since they are more focused in their search. However, when the heuristic values are poorly informed (small weights), they explore more unique contexts. Agents of ADOPT explore many more repeated contexts than agents of BnB-ADOPT since they need to reconstruct partial solutions that they purged from memory. Agents of BnB-ADOPT explore a few repeated contexts even though it does not have to reconstruct partial solutions. The reason for this behavior is the distributed nature of BnB-ADOPT. For example, assume that the context of an agent is $\{(a_1, 0), (a_2, 0)\}$ and the next context of a centralized variant of BnB-ADOPT would be $\{(a_1, 1), (a_2, 1)\}$ (where the IDs are omitted for simplicity). The agent updates its context to $\{(a_1, 1), (a_2, 0)\}$ when it receives the message from agent $a_1$ that it takes on value 1. The agent then updates its context to $\{(a_1, 1), (a_2, 1)\}$ when it receives the message from agent $a_2$ that it takes on value 1. Thus, the agent explores the intermediate context $\{(a_1, 1), (a_2, 0)\}$ that a centralized variant of BnB-ADOPT would not explore. It counts as a repeated context if the agent explores this context intentionally in the future. Overall, BnB-ADOPT tends to be faster than ADOPT if the heuristic values are poorly informed (small weights). Thus, BnB-ADOPT has great potential as a DCOP search algorithm since heuristic values are often poorly informed for complex DCOP problems, such as DCOP problems with large numbers of agents, large domains, large numbers of constraints or large ranges of constraint costs.

## 6.4 Experimental Results: Suboptimal Variants of BnB-ADOPT

We now compare the three suboptimal variants of BnB-ADOPT to each other. The experimental setup is identical to the one for the optimal DCOP search algorithms, except as follows: For graph coloring problems, the number of vertices is 10, the range of constraint costs is 0 to 10,000 and the constraint density is 2; for sensor network problems, the number of targets is 9; and for meeting scheduling problems, the number of meetings is 10. We measure the runtimes in cycles. (The results for NCCCs are similar.) However, we report normalized runtimes, that is, the runtimes divided by the runtime for finding a cost-minimal solution with BnB-ADOPT. Thus, the normalized runtime 0.25 refers to one quarter of the number of cycles that it takes to find a cost-minimal solution with BnB-ADOPT. Similarly, we report normalized solution costs, that is, the solution costs divided by the minimal solution costs. Thus, the normalized solution cost 2.5 refers to a solution cost that is two and a half times larger than the minimal solution cost. We vary the relative error bound (which is the worst acceptable normalized solution cost) from 1.0 to 4.0. The relative error bound is $p$ for BnB-ADOPT$_{REM}$ and $w$ for BnB-ADOPT$_{WHM}$. We pre-calculate the minimal solution costs to set the correct value of $b$ for BnB-ADOPT$_{AEM}$. For example, if the minimal solution cost is 100 and the relative error bound is 2.5, then $p = 2.5$ for BnB-ADOPT$_{REM}$, $w = 2.5$ for BnB-ADOPT$_{WHM}$ and $b = (2.5 - 1) \cdot 100 = 150$ for BnB-ADOPT$_{AEM}$.

Figure 21(a-c) shows our experimental results for graph coloring problems. Figure 21(a) shows that the normalized solution costs of all three suboptimal variants increase as the relative error bound increases. However, the solution costs remain much smaller than the error bound. For example, the normalized solution costs of all three suboptimal variants are less than 1.3 (rather than 3) when the relative error bound is 3. The normalized solution costs of BnB-ADOPT$_{AEM}$ are usually larger than the normalized solution costs of BnB-ADOPT$_{REM}$ for the same relative error bound. The reason for this behavior is that BnB-ADOPT$_{AEM}$ terminates when $UB_{X^r}^r \leq lim^r = b + LB_{X^r}^r = (p-1) \cdot \gamma_{X^r}^r + LB_{X^r}^r$, where $\gamma_{X^r}^r$ is the minimal solution cost. Thus, the solution cost of BnB-ADOPT$_{AEM}$ can be at most $UB_{X^r}^r - LB_{X^r}^r \leq (p-1) \cdot \gamma_{X^r}^r$ larger than the minimal solution cost. On the other hand, BnB-ADOPT$_{REM}$ terminates when $UB_{X^r}^r \leq lim^r = p \cdot LB_{X^r}^r$. Thus, the solution cost of BnB-ADOPT$_{REM}$ can be at most $UB_{X^r}^r - LB_{X^r}^r \leq (p-1) \cdot LB_{X^r}^r$ larger than the minimal solution cost. The absolute error bound of BnB-ADOPT$_{AEM}$ is thus no smaller than the absolute error bound of BnB-ADOPT$_{REM}$ since $\gamma_{X^r}^r \geq LB_{X^r}^r$ but is initially strictly greater than the absolute error bound of BnB-ADOPT$_{REM}$ since $\gamma_{X^r}^r > LB_{X^r}^r$ during most of the search.

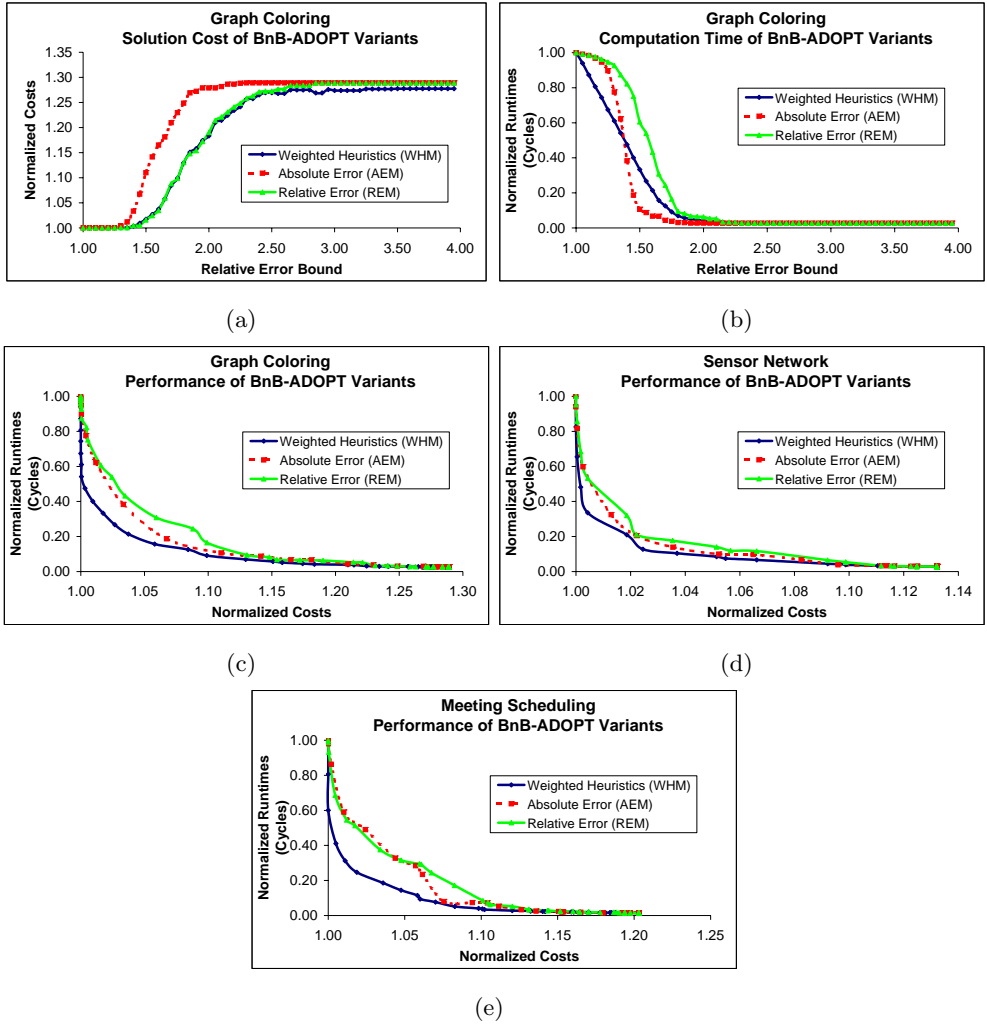

Figure 21: Experimental Results Comparing the Suboptimal Variants of BnB-ADOPT

Figure 21(b) shows that the normalized runtimes of all three suboptimal variants decrease as the relative error bound increases. They decrease to almost 0 when the relative error bound is about 2.0. Therefore, all three suboptimal variants terminate almost immediately after finding the first solution. The normalized runtimes of BnB-ADOPT$_{AEM}$ are usually smaller than the normalized runtimes of BnB-ADOPT$_{REM}$ for the same relative error bound since BnB-ADOPT$_{AEM}$ can terminate with a suboptimal solution cost that is within its absolute error bound but not yet within the absolute error bound of BnB-ADOPT$_{REM}$ if the absolute error bound of BnB-ADOPT$_{AEM}$ is strictly greater than the absolute error bound of BnB-ADOPT$_{REM}$. In other words, BnB-ADOPT$_{AEM}$ can terminate with a suboptimal solution cost $(p-1) \cdot LB_{X^r}^r < UB_{X^r}^r \leq (p-1) \cdot \gamma_{X^r}^r$ while BnB-ADOPT$_{REM}$ can not.

Figure 21(c) shows the normalized runtimes needed to achieve a given normalized solution cost. BnB-ADOPT$_{WHM}$ terminates faster than BnB-ADOPT$_{AEM}$, which in turn terminates faster than BnB-ADOPT$_{REM}$. For example, the normalized runtime needed to achieve the normalized solu-tion cost 1.05 is about 0.18 for BnB-ADOPT$_{WHM}$, 0.30 for BnB-ADOPT$_{AEM}$ and 0.35 for BnB-ADOPT$_{REM}$. Thus, BnB-ADOPT$_{WHM}$ is the suboptimal variant of BnB-ADOPT with the best performance. Figures 21(d-e) show the same trend for sensor network and meeting scheduling problems.

## 7. Conclusions

In this article, we introduced Branch-and-Bound ADOPT (BnB-ADOPT), a memory-bounded DCOP search algorithm. BnB-ADOPT uses the message passing and communication framework of ADOPT but changes the search strategy of ADOPT from best-first search to depth-first branch-and-bound search to make ADOPT faster by taking advantage of the fact that DCOP problems have depth-bounded search trees. The other properties of BnB-ADOPT are similar to those of ADOPT. BnB-ADOPT allows agents to operate concurrently (in order to decrease the runtime) and asynchronously (in order to increase robustness). BnB-ADOPT restricts communication to agents that share constraints (in order to fit the restrictions of applications such as sensor networks). Finally, BnB-ADOPT orders agents into a pseudo-tree (in order to take advantage of independent DCOP subproblems). Our experimental results showed that BnB-ADOPT finds cost-minimal solutions up to one order of magnitude faster than ADOPT for a variety of large DCOP problems and is as fast as NCBB for most of these DCOP problems. The reason for this behavior is the following: Agents of NCBB operate sequentially and are thus often idle. ADOPT can construct fewer partial solutions than BnB-ADOPT but has to reconstruct some partial solutions that it purged from memory. The advantage of ADOPT with respect to the number of constructed partial solutions decreases and its disadvantage with respect to the number of reconstructed partial solutions increases as heuristic values become more poorly informed. Thus, BnB-ADOPT has great potential as a DCOP search algorithm since heuristic values are often poorly informed for complex DCOP problems such as DCOP problems with large numbers of agents, large domains, large numbers of constraints or large ranges of constraint costs.

We also investigated three approximation mechanisms that trade off the solution cost of BnB-ADOPT for a smaller runtime, namely the Absolute Error Mechanism from ADOPT (resulting in BnB-ADOPT$_{AEM}$), the new Relative Error Mechanism (resulting in BnB-ADOPT$_{REM}$) and the new Weighted Heuristics Mechanism (resulting in BnB-ADOPT$_{WHM}$). The two new approximation mechanisms allow users to specify a relative error bound, which is often more meaningful than an absolute error bound. The Weighted Heuristics Mechanism dominated both the Absolute Error Mechanism and the Relative Error Mechanism in our experiments and should apply to other DCOP search algorithms as well since they all benefit from using heuristic values to focus their searches (Yeoh, Koenig, & Sun, 2008b).

In the future, we plan to improve BnB-ADOPT in the following ways: First, we would like to reduce the number of sent messages and handle lost messages. Second, we would like to study how different pseudo-tree arrangements (Atlas & Decker, 2007; Sultanik, Lass, & Regli, 2009) and pre-processing techniques (Matsui et al., 2009) affect the efficiency of BnB-ADOPT. Finally, we would like to compare BnB-ADOPT and its approximation mechanisms to other DCOP algorithms, including OptAPO, DPOP and their variants (Petcu & Faltings, 2005a, 2006).

## Acknowledgments

This article is an extension of two earlier publications (Yeoh, Felner, & Koenig, 2008a; Yeoh et al., 2008b) and contains additional expositions, examples and proofs. We thank Anton Chechetka for providing us with his implementation of NCBB and the anonymous reviewers for their helpful comments. This research was done while Ariel Felner spent his sabbatical at the University of Southern California, visiting Sven Koenig. This research has been partly supported by a U.S. Army

Research Laboratory (ARL) and U.S. Army Research Office (ARO) award to Sven Koenig under grant W911NF-08-1-0468, by a Office of Naval Research (ONR) award to Sven Koenig under grant N00014-09-1-1031, by a National Science Foundation (NSF) award to Sven Koenig under grant 0413196 and by an Israeli Science Foundation (ISF) award to Ariel Felner under grants 728/06 and 305/09. The views and conclusions contained in this document are those of the authors and should not be interpreted as representing the official policies, either expressed or implied, of the sponsoring organizations, agencies, companies or the U.S. government.

# References

Ali, S., Koenig, S., & Tambe, M. (2005). Preprocessing techniques for accelerating the DCOP algorithm ADOPT. In *Proceedings of the International Joint Conference on Autonomous Agents and Multiagent Systems (AAMAS)*, pp. 1041–1048.

Atlas, J., & Decker, K. (2007). A complete distributed constraint optimization method for non-traditional pseudotree arrangements. In *Proceedings of the International Joint Conference on Autonomous Agents and Multiagent Systems (AAMAS)*, pp. 736–743.

Bayardo, R., & Miranker, D. (1995). On the space-time trade-off in solving constraint satisfaction problems. In *Proceedings of the International Joint Conference on Artificial Intelligence (IJCAI)*, pp. 558–562.

Bistarelli, S., Montanari, U., Rossi, F., Schiex, T., Verfaillie, G., & Fargier, H. (1999). Semiring-based CSPs and valued CSPs: Basic properties and comparison. *Constraints*, *4*(3), 199–240.

Bowring, E., Pearce, J., Portway, C., Jain, M., & Tambe, M. (2008). On k-optimal distributed constraint optimization algorithms: New bounds and algorithms. In *Proceedings of the International Joint Conference on Autonomous Agents and Multiagent Systems (AAMAS)*, pp. 607–614.

Bowring, E., Tambe, M., & Yokoo, M. (2006). Multiply-constrained distributed constraint optimization. In *Proceedings of the International Joint Conference on Autonomous Agents and Multiagent Systems (AAMAS)*, pp. 1413–1420.

Burke, D., & Brown, K. (2006). Efficiently handling complex local problems in distributed constraint optimisation. In *Proceedings of the European Conference on Artificial Intelligence (ECAI)*, pp. 701–702.

Chechetka, A., & Sycara, K. (2006). No-commitment branch and bound search for distributed constraint optimization. In *Proceedings of the International Joint Conference on Autonomous Agents and Multiagent Systems (AAMAS)*, pp. 1427–1429.

Choxi, H., & Modi, P. (2007). A distributed constraint optimization approach to wireless network optimization. In *Proceedings of the AAAI-07 Workshop on Configuration*, pp. 1–8.

Davin, J., & Modi, P. (2006). Hierarchical variable ordering for multiagent agreement problems. In *Proceedings of the International Joint Conference on Autonomous Agents and Multiagent Systems (AAMAS)*, pp. 1433–1435.

Dechter, R. (Ed.). (2003). *Constraint Processing*. Morgan Kaufmann.

Fitzpatrick, S., & Meertens, L. (2003). Distributed coordination through anarchic optimization. In Lesser, V., Ortiz, C., & Tambe, M. (Eds.), *Distributed Sensor Networks: A Multiagent Perspective*, pp. 257–295. Kluwer.

Freuder, E., & Quinn, M. (1985). Taking advantage of stable sets of variables in constraint satisfaction problems. In *Proceedings of the International Joint Conference on Artificial Intelligence (IJCAI)*, pp. 1076–1078.

Gershman, A., Meisels, A., & Zivan, R. (2009). Asynchronous Forward-Bounding for distributed COPs. *Journal of Artificial Intelligence Research, 34*, 61–88.

Greenstadt, R. (2009). An overview of privacy improvements to k-optimal DCOP algorithms (extended abstract). In *Proceedings of the International Joint Conference on Autonomous Agents and Multiagent Systems (AAMAS)*, pp. 1279–1280.

Greenstadt, R., Grosz, B., & Smith, M. (2007). SSDPOP: Improving the privacy of DCOP with secret sharing. In *Proceedings of the International Joint Conference on Autonomous Agents and Multiagent Systems (AAMAS)*, pp. 1098–1100.

Hamadi, Y., Bessière, C., & Quinqueton, J. (1998). Distributed intelligent backtracking. In *Proceedings of the European Conference on Artificial Intelligence (ECAI)*, pp. 219–223.

Hirayama, K., & Yokoo, M. (1997). Distributed partial constraint satisfaction problem. In *Proceedings of the International Conference on Principles and Practice of Constraint Programming (CP)*, pp. 222–236.

Jain, M., Taylor, M., Tambe, M., & Yokoo, M. (2009). DCOPs meet the real world: Exploring unknown reward matrices with applications to mobile sensor networks. In *Proceedings of the International Joint Conference on Artificial Intelligence (IJCAI)*, pp. 181–186.

Junges, R., & Bazzan, A. (2008). Evaluating the performance of DCOP algorithms in a real world, dynamic problem. In *Proceedings of the International Conference on Autonomous Agents and Multiagent Systems (AAMAS)*, pp. 599–606.

Korf, R. (1993). Linear-space best-first search. *Artificial Intelligence, 62*(1), 41–78.

Kumar, A., Faltings, B., & Petcu, A. (2009). Distributed constraint optimization with structured resource constraints. In *Proceedings of the International Joint Conference on Autonomous Agents and Multiagent Systems (AAMAS)*, pp. 923–930.

Lesser, V., Ortiz, C., & Tambe, M. (Eds.). (2003). *Distributed Sensor Networks: A Multiagent Perspective*. Kluwer.

Maheswaran, R., Pearce, J., & Tambe, M. (2004a). Distributed algorithms for DCOP: A graphical game-based approach. In *Proceedings of the International Conference on Parallel and Distributed Computing Systems (PDCS)*, pp. 432–439.

Maheswaran, R., Tambe, M., Bowring, E., Pearce, J., & Varakantham, P. (2004b). Taking DCOP to the real world: Efficient complete solutions for distributed event scheduling. In *Proceedings of the International Joint Conference on Autonomous Agents and Multiagent Systems (AAMAS)*, pp. 310–317.

Mailler, R., & Lesser, V. (2004). Solving distributed constraint optimization problems using cooperative mediation. In *Proceedings of the International Joint Conference on Autonomous Agents and Multiagent Systems (AAMAS)*, pp. 438–445.

Marinescu, R., & Dechter, R. (2007). Best-first AND/OR search for graphical models. In *Proceedings of the AAAI Conference on Artificial Intelligence (AAAI)*, pp. 1171–1176.

Marinescu, R., & Dechter, R. (2009). AND/OR branch-and-bound search for combinatorial optimization in graphical models. *Artificial Intelligence, 173*(16-17), 1457–1491.

Matsui, T., Silaghi, M., Hirayama, K., Yokoo, M., & Matsuo, H. (2009). Directed soft arc consistency in pseudo trees. In *Proceedings of the International Conference on Autonomous Agents and Multiagent Systems (AAMAS)*, pp. 1065–1072.

Meisels, A., Kaplansky, E., Razgon, I., & Zivan, R. (2002). Comparing performance of distributed constraints processing algorithms. In *Proceedings of the Distributed Constraint Reasoning Workshop*, pp. 86–93.

Modi, P., & Ali, S. (2004). Distributed constraint reasoning under unreliable communication. In Zhang, W., & Sorge, V. (Eds.), *Frontiers in Artificial Intelligence and Applications*, Vol. 112, pp. 141–150. IOS Press.

Modi, P., Shen, W.-M., Tambe, M., & Yokoo, M. (2005). ADOPT: Asynchronous distributed constraint optimization with quality guarantees. *Artificial Intelligence*, *161*(1-2), 149–180.

Ottens, B., & Faltings, B. (2008). Coordinating agent plans through distributed constraint optimization. In *Proceedings of the ICAPS-08 Workshop on Multiagent Planning*.

Pearce, J., & Tambe, M. (2007). Quality guarantees on k-optimal solutions for distributed constraint optimization problems. In *Proceedings of the International Joint Conference on Artificial Intelligence (IJCAI)*, pp. 1446–1451.

Pecora, F., Modi, P., & Scerri, P. (2006). Reasoning about and dynamically posting n-ary constraints in ADOPT. In *Proceedings of the Distributed Constraint Reasoning Workshop*, pp. 57–71.

Petcu, A., & Faltings, B. (2005a). Approximations in distributed optimization. In *Proceedings of the International Conference on Principles and Practice of Constraint Programming (CP)*, pp. 802–806.

Petcu, A., & Faltings, B. (2005b). A scalable method for multiagent constraint optimization. In *Proceedings of the International Joint Conference on Artificial Intelligence (IJCAI)*, pp. 1413–1420.

Petcu, A., & Faltings, B. (2006). ODPOP: An algorithm for open/distributed constraint optimization. In *Proceedings of the National Conference on Artificial Intelligence (AAAI)*, pp. 703–708.

Pohl, I. (1970). First results on the effect of error in heuristic search. *Machine Intelligence*, *5*, 219–236.

Pohl, I. (1973). The avoidance of (relative) catastrophe, heuristic competence, genuine dynamic weighting and computational issues in heuristic problem solving. In *Proceedings of the International Joint Conference on Artificial Intelligence (IJCAI)*, pp. 12–17.

Schiex, T., Fargier, H., & Verfaillie, G. (1995). Valued constraint satisfaction problems: Hard and easy problems. In *Proceedings of the International Joint Conference on Artificial Intelligence (IJCAI)*, pp. 631–637.

Schurr, N., Okamoto, S., Maheswaran, R., Scerri, P., & Tambe, M. (2005). Evolution of a teamwork model. In Sun, R. (Ed.), *Cognition and Multi-Agent Interaction: From Cognitive Modeling to Social Simulation*, pp. 307–327. Cambridge University Press.

Silaghi, M., Landwehr, J., & Larrosa, J. (2004). Asynchronous branch & bound and A* for disWCSPs with heuristic function based on consistency-maintenance. In Zhang, W., & Sorge, V. (Eds.), *Frontiers in Artificial Intelligence and Applications*, Vol. 112, pp. 49–62. IOS Press.

Silaghi, M., Lass, R., Sultanik, E., Regli, W., Matsui, T., & Yokoo, M. (2008). The operation point units of distributed constraint solvers. In *Proceedings of the Distributed Constraint Reasoning Workshop*, pp. 1–16.

Silaghi, M., & Yokoo, M. (2009). ADOPT-ing: Unifying asynchronous distributed optimization with asynchronous backtracking. *Autonomous Agents and Multi-Agent Systems*, *19*(2), 89–123.

Stranders, R., Farinelli, A., Rogers, A., & Jennings, N. (2009). Decentralised coordination of mobile sensors using the Max-Sum algorithm. In *Proceedings of the International Joint Conference on Artificial Intelligence (IJCAI)*, pp. 299–304.

Sultanik, E., Lass, R., & Regli, W. (2009). Dynamic configuration of agent organizations. In *Proceedings of the International Joint Conference on Artificial Intelligence (IJCAI)*, pp. 305–311.

Yeoh, W., Felner, A., & Koenig, S. (2008a). BnB-ADOPT: An asynchronous branch-and-bound DCOP algorithm. In *Proceedings of the International Joint Conference on Autonomous Agents and Multiagent Systems (AAMAS)*, pp. 591–598.

Yeoh, W., Koenig, S., & Sun, X. (2008b). Trading off solution cost for smaller runtime in DCOP search algorithms (short paper). In *Proceedings of the International Joint Conference on Autonomous Agents and Multiagent Systems (AAMAS)*, pp. 1445–1448.

Yeoh, W., Varakantham, P., & Koenig, S. (2009). Caching schemes for DCOP search algorithms. In *Proceedings of the International Joint Conference on Autonomous Agents and Multiagent Systems (AAMAS)*, pp. 609–616.

Yokoo, M., & Hirayama, K. (1996). Distributed breakout algorithm for solving distributed constraint satisfaction problems. In *Proceedings of the International Conference on Multiagent Systems (ICMAS)*, pp. 401–408.

Zhang, W., & Korf, R. (1995). Performance of linear-space search algorithms. *Artificial Intelligence*, *79*(2), 241–292.

Zhang, W., Xing, Z., Wang, G., & Wittenburg, L. (2003). An analysis and application of distributed constraint satisfaction and optimization algorithms in sensor networks. In *Proceedings of the International Joint Conference on Autonomous Agents and Multiagent Systems (AAMAS)*, pp. 185–192.

Zivan, R. (2008). Anytime local search for distributed constraint optimization. In *Proceedings of the AAAI Conference on Artificial Intelligence (AAAI)*, pp. 393–398.

Zivan, R., Glinton, R., & Sycara, K. (2009). Distributed constraint optimization for large teams of mobile sensing agents. In *Proceedings of the International Conference on Intelligent Agent Technology (IAT)*, pp. 347–354.